\documentclass[lettersize,journal]{IEEEtran}
\usepackage{titlesec}
\titleformat{\subsection}
  {\normalfont\bfseries\itshape}
  {}
  {0pt}
  {\Alph{subsection}.\ }

\titleformat*{\subsubsection}{\bfseries\itshape}

\usepackage{amsmath,amsfonts}
\usepackage{array}
\usepackage[caption=false,font=normalsize,labelfont=sf,textfont=sf]{subfig}

\usepackage{textcomp}
\usepackage{stfloats}
\usepackage{url}
\usepackage{verbatim}
\usepackage{graphicx}
\usepackage{cite}
\usepackage{diagbox}
\usepackage{makecell}
\usepackage{color}
\definecolor{brickred}{rgb}{0.8, 0.25, 0.33}
\definecolor{brickred2}{rgb}{0.25, 0.8, 0.33}
\usepackage[linesnumbered,ruled,lined,commentsnumbered]{algorithm2e}
\usepackage[table,xcdraw]{xcolor}
\usepackage{color}
\usepackage{booktabs}
\usepackage{multirow}
\usepackage{adjustbox}
\usepackage{animate}
\usepackage{listings}
\usepackage{microtype}
\usepackage[breakable]{tcolorbox}
\newtcolorbox{mybox}{fontupper=\scriptsize\tt}

\newcommand{\etal}{{\it et al.}}
\newcommand{\eg}{{\it e.g.}}
\newcommand{\ie}{{\it i.e.}}

\begin{document}


\title{Harnessing Meta-Learning for\\ Controllable Full-Frame Video Stabilization}


\author{Muhammad~Kashif~Ali, 
        Eun~Woo~Im,
        Dongjin~Kim,
        Tae~Hyun~Kim,
        Vivek~Gupta,
        Haonan~Luo,
        Tianrui~Li%

\IEEEcompsocitemizethanks{\IEEEcompsocthanksitem
  M. K. Ali, H. Luo, and T. Li are with the School of Computing and Artificial Intelligence, Southwest Jiaotong University, China. 
  
  D. Kim and T. H. Kim are with the Department of Computer Science, Hanynag University, South Korea. 
  
  E. W. Im and V. Gupta are with the School of Computing and Augmented Intelligence, Arizona State University, USA.\\
 (Corresponding authors: Tae Hyun Kim, Haonan Luo, Tianrui Li)
 \protect\\

}
}


\markboth{Journal of \LaTeX\ Class Files,~Vol.~14, No.~8, August~2021}%
{Shell \MakeLowercase{\textit{et al.}}: A Sample Article Using IEEEtran.cls for IEEE Journals}

\IEEEpubid{0000--0000/00\$00.00~\copyright~2021 IEEE}

\maketitle

\begin{abstract}
Video stabilization remains a fundamental problem in computer vision, particularly pixel-level synthesis solutions for video stabilization, which synthesize full-frame outputs, add to the complexity of this task. These methods aim to enhance stability while synthesizing full-frame videos, but the inherent diversity in motion profiles and visual content present in each video sequence makes robust generalization with fixed parameters difficult. 
To address this, we present a novel method that improves pixel-level synthesis video stabilization methods by rapidly adapting models to each input video at test time. The proposed approach takes advantage of low-level visual cues available during inference to improve both the stability and visual quality of the output. 
Notably, the proposed rapid adaptation achieves significant performance gains even with a single adaptation pass.
We further propose a jerk localization module and a targeted adaptation strategy, which focuses the adaptation on high-jerk segments for maximizing stability with fewer adaptation steps. 
The proposed methodology enables modern stabilizers to overcome the longstanding SOTA approaches while maintaining the full frame nature of the modern methods, while offering users with control mechanisms akin to classical approaches.
Extensive experiments on diverse real-world datasets demonstrate the versatility of the proposed method. Our approach consistently improves the performance of various full-frame synthesis models in both qualitative and quantitative terms, including results on downstream applications.

\end{abstract}


\section{Introduction}

\IEEEPARstart{R}{ecording} and sharing videos has recently become a standard practice in today's life, as millions of individuals utilize platforms like YouTube and Facebook to document and share memorable moments from their daily lives.
However, the absence of specialized stabilization equipment, such as gimbals, often results in shaky videos.
This shakiness degrades the viewing experience and hampers visual communication.
Consequently, video stabilization has attracted significant attention from both academic researchers and industry practitioners alike, given its potential to enhance visual quality and improve downstream computer vision tasks such as tracking, segmentation, and semantic understanding.

Conventionally, video stabilization approaches have relied on a straightforward pipeline comprising of motion estimation, smoothing, and motion compensation techniques involving spatial transformations~\cite{grundmann2011auto, liu2013bundled, yu2020learning}. Despite significant strides in improving these components, these methodologies result in a loss of valuable visual content due to pixel transformations, leading to irregular boundaries near the edges of the stabilized videos. To mitigate these boundary distortions, simple center cropping is widely employed to conceal irregular boundaries at the cost of visual resolution. 
However, recent deep learning methodologies present promising alternative strategies to minimize the loss of visual resolution. 
Techniques such as inpainting missing regions~\cite{OVS, choi2021self} and joint stabilization-synthesis pipelines~\cite{bmvc_ver, DIF, liu2021hybrid} have achieved notable success. However, end-to-end pixel-level regressive stabilization remains a challenging task due to the inherent complexity of this task and the diversity of real-world motion scenarios hindering robust generalization.
\IEEEpubidadjcol


However, pioneering works such as DIFRINT~\cite{DIF} and DMBVS~\cite{bmvc_ver} represent significant strides toward an end-to-end full-frame stabilization solution. DIFRINT achieves stabilization through recursive temporal interpolation guided by optical flow, while DMBVS introduced a large-scale dataset of paired stable and unstable videos with similar perspectives and proposed a direct pixel-synthesis approach for regressing stabilized videos. 
Despite their success, both methods exhibit certain limitations. DMBVS, while producing high-quality full-frame videos, lacks any mechanism to control the level of stability. Whereas, DIFRINT struggles with maintaining perceptual quality due to repeated frame interpolation iterations and is susceptible to temporal artifacts, particularly near motion boundaries involving occlusion and disocclusion.

To address these challenges, we propose an adaptive stabilization framework that utilizes meta-learning to enable rapid adaptation of these stabilization models at test time. We hypothesize that meta-trained models can quickly adjust to input scene and motion cues during inference. This adaptability allows the models to handle a wide range of motion patterns while also offering users with a control mechanism, a feature traditionally limited to classical transformation-based methodologies. By incorporating this control mechanism, the proposed approach effectively bridges the gap between modern learning-based and classical transformation-based techniques, delivering high-quality full-frame stabilized videos with controllable stability.

The proposed framework performs adaptation using only unstable inputs, adjusting the model parameters to the unique content and dynamics of each video. We demonstrate that even a single adaptation pass yields substantial improvements in stability (up to $\sim$8\% absolute gain) on the standard NUS benchmark dataset~\cite{liu2013bundled}, surpassing the longstanding state-of-the-art (SOTA) methodologies.

As an additional contribution to our prior work~\cite{meta_stab}, we introduce an efficient motion-guided adaptation strategy via jerk localization and spatially aware adaptation, which further expedites the adaptation process. The proposed jerk localization module identifies regions of high-intensity jerks, enabling the model to focus adaptation efforts where they are most needed, thereby maximizing stability gains with minimum adaptation overhead. This targeted strategy further highlights the controllability of the proposed approach, and its expedient nature makes it highly suitable for real-world deployment for professional applications.

To broaden the scope of the evaluation, we further assess the proposed method in challenging real-world data sets such as BiT~\cite{BiT} and DOFVS~\cite{DOFVS}, where it consistently achieves SOTA results. To evaluate the real-world application of video stabilization approaches in modern vision systems, we introduce two novel metrics to both quantify the performance stabilization approaches and to assess their utility in vision-based systems. More specifically, we introduce metrics to assess the tracking feasibility and coherence across frames. 

Additionally, we evaluate the impact of video stabilization on downstream tasks in modern vision systems, focusing on Large Video-Language Models (LViLMs). To this end, we introduce an automated LLM-as-a-Judge framework for scalable assessment of high-level video semantic understanding, offering both a cost-effective alternative to resource-intensive user studies and a method to highlight the viability of video stabilization in modern vision pipelines.
These evaluations are conducted comprehensively at scale to rigorously assess the performance of various stabilization techniques, including professional, classical, and modern methods. Our findings suggest that improved stability and visual quality not only enhance human viewing experience but also significantly benefit machine-level video understanding.

In this work, we expand our previous work~\cite{meta_stab}, and significantly improve the methodology, experiments, and analysis. We summarize our contributions as follows.
\begin{itemize}
  \item We propose an adaptive stabilization framework based on meta-learning, enabling full-frame pixel-synthesis models to adapt using only unstable input videos.
  \item We introduce a jerk localization module and a spatially targeted adaptation strategy that guide faster adaptation to temporally unstable regions for efficient motion-aware stabilization and achieve competitive performance with minimal adaptation overhead.
  \item We propose and evaluate two novel metrics to better assess the impact of stabilization on downstream vision tasks and their potential use in vision-based systems.
  \item We also introduce a scalable evaluation protocol using Large Video-Language Models (LViLMs) as an alternative to tedious user studies, demonstrating the effectiveness of our method and highlighting the viability of stabilization in downstream applications and modern vision systems.

  
\end{itemize}

\section{Related works}
\subsection{Video stabilization}
Traditionally, video stabilization methodologies can be classified into three distinct categories: 3D, 2.5D, and 2D approaches.
The 3D video stabilization approaches model camera trajectories in the 3D space. Various techniques such as depth information~\cite{liu2012video}, gyroscopic sensor data~\cite{karpenko2011digital, DOFVS, DUT} structure from motion (SfM)~\cite{liu2009content}, light fields~\cite{smith2009light}, and 3D plane constraints~\cite{zhou2013plane} have been investigated to stabilize videos. Apart from these classical approaches, modern 3D video stabilization approaches have also investigated hybrid fusion methodologies for producing full-frame videos~\cite{liu2021hybrid}. 

Despite their innovative techniques, these methodologies face difficulties in handling dynamic scenes containing multiple moving objects and require extensive computational resources; therefore, 2D approaches, which limit their scope to spatial transformations, homography, and affine transformations became the tool of choice for researchers and professionals alike.
Conventionally, these approaches track and stabilize the trajectories of prominent features. Doing so introduces severe loss of visual content near the frame boundaries, which is often concealed by upscaling the resultant frames and center-cropping.

For 2D stabilization, Buehler~\etal~\cite{buehler2001non} estimated camera poses in shaky videos and used non-metric image-based rendering to stabilize videos.
Matsushita~\etal~\cite{matsushita2006full} introduced simplistic 2D global transformations to warp the unstable frames to produce stable video, and Liu~\etal~\cite{liu2013bundled} extended this phenomenon to grid-based warping for smoothing feature trajectories.
Grundmann~\etal~\cite{grundmann2011auto} presented an $L_1$-based objective function for estimating stable camera trajectories, whereas Liu~\etal~\cite{liu2011subspace} utilized eigen-trajectory smoothing for this task.
Goldstein~\etal~\cite{goldstein2012video}, Lee~\etal~\cite{lee2009video}, and Wang~\etal~\cite{wang2013spatially} employed epipolar geometry-based optimization models for stabilizing videos.

Inspired by these approaches and looking at the limitation of these approaches in handling the independent motion of multiple objects, Liu~\etal~\cite{liu2013bundled} highlighted the importance of ``\emph{relatively}" denser inter-frame motion through optical flow for video stabilization.
Their findings inspired most of the modern video stabilization methodologies that are currently being used professionally to this day in apps like Blink, Adobe Premiere Pro, and Deshaker. 
Many recent works~\cite{OVS, DUT, DIF, yu2019robust, yu2020learning, liu2021hybrid} rely on dense optical flow as an irreplaceable backbone in their approaches. Geo et al.\cite{geo2023globalflownet} further improved on these methods and finetuned a conventional flow estimation network to estimate only the camera motion component of optical flow (termed global optical flow) and used it to define warping fields for video stabilization.

Please note that, unlike the conventional methodologies for deep stabilization, Ali~\etal~\cite{bmvc_ver} highlighted the importance of perspective in training data and the power of traditional deep convolutional neural networks (CNNs) by learning to synthesize stable frames entirely through learned implicit motion compensation from neighboring frames, and Choi~\etal~\cite{DIF} proposed an iterative interpolation strategy for stabilizing videos.
Please note that these two methods are the only proposed methods for pixel synthesis end-to-end solutions for full-frame video stabilization.

\subsection{Meta learning and test-time optimization}
For deep video stabilization methods, some literature has been investigated on test-time adaptation inspired by the conventional optimization approaches.
Yu~\etal~\cite{yu2019robust} proposed to stabilize videos by optimizing the motion vector warp field in CNN weight-space, and Xu~\etal~\cite{DUT} defined a pipeline inspired by~\cite{grundmann2011auto, liu2013bundled} with the help of a modular pipeline catering to estimating and iteratively smoothing the motion trajectories and reprojecting the unstable frames to follow a smooth global motion profile. Despite the ingenuity of these approaches, these methods require significant computational resources and time for stabilizing videos. 

Contrary to the conventional test-time optimization-based video stabilization approaches, we aim to investigate faster test-time adaptability for full-frame video stabilization approaches inspired by its recent success in various computer vision tasks such as super-resolution~\cite{park2020fast, gupta2021ada, lee2021dynavsr}, visual tracking~\cite{metatrack}, video segmentation~\cite{behl2020meta}, object detection~\cite{deng2021minet}, human pose estimation~\cite{cho2021camera}, image enhancement~\cite{ma2023bilevel}, and video frame interpolation~\cite{metavfi}.

Conventionally, meta-learning algorithms can be categorized into three main groups: metric-based, network-based, and optimization-based algorithms.
Metric-based algorithms are known for encoding prior knowledge by acquiring a feature embedding space~\cite{metric_based_1, metric_based_2, metric_based_3}.
On the other hand, network-based meta-learning methodologies often leverage prior knowledge into the architecture of the considered models~\cite{network_based_1, network_based_2, network_based_3}.
Despite their respective strengths, both metric and network-based systems have certain limitations in terms of their applications or scalability.
In contrast, optimization-based methodologies overcome these limitations by adjusting the optimization algorithm during the adaptation process.
This characteristic enhances their potential to effectively accommodate a wider range of applications.

From the optimization-based category of meta-learning, model-agnostic meta-learning (MAML)~\cite{maml} has become the go-to method for researchers investigating various computer vision tasks~\cite{cheng2021meta, fu2019embodied, hosseinzadeh2023few, lee2019metapix, lin2021self, lu2020deep, ren2020video, wang2021meta, wang2020tracking, yang2023toward, zhao2021one, zou20202, min2023meta} due to its flexibility, effectiveness and generalizability.

In light of recent advancements and its success in low-level computer vision tasks, we investigate the applicability of this technique for improving the performance of pixel-level synthesis solutions for video stabilization and propose a new algorithm that combines the strengths of conventional spatial transformation-guided video stabilization approaches with the full-frame nature of regressive pixel-level synthesis video stabilization approaches.
The proposed algorithm allows the parameters of the considered video stabilization models to be updated quickly with respect to the unique motion profiles and diverse visual content present in each video, and allows the adapted model to stabilize videos while preserving visual quality and resolution.
The proposed algorithm additionally allows the user to adjust the stability and maintain quality  according to their preference (to a certain extent), which was not possible with the existing regressive solutions for this task.

\section{Proposed method}

This section covers pixel-level regressive video stabilization by providing an in-depth description of the proposed algorithms, meta-training objectives, and comprehensive network architectures, in addition to discussing several inference strategies for this task.

\subsection{Problem set-up}
Consider an unstable video, which comprises consecutive video frames as $V$ = \{$I_{0}$, $I_{1}$, \dots , $I_{T}$\}. 
The goal of video stabilization methods is to produce a stable video $\hat{V}$ = \{$\hat{I}_{0}$, $\hat{I}_{1}$, \dots, $\hat{I}_{T}$\} using a stabilization network $f_{\theta}$ given unstable input video $V$. The output video $\hat{V}$ contains content similar to $V$ with a stabilized camera trajectory.
Conventionally, stabilization techniques based on pixel synthesis~\cite{bmvc_ver, DIF} typically adopt a sliding window protocol. This approach processes a local temporal window that includes $2k + 1$ consecutive frames at a given time step $t$, represented by ${S}_t=\{I_{t - k}, ..., I_{t}, ..., I_{t + k}\}$. 
The stabilized frame $\hat{I_{t}}$ is then regressed as follows:
\begin{equation}
    \hat{I_{t}} = f_{\theta}({S}_t).
\label{eq:frame_ip_op_non_rec}
\end{equation}
This local temporal window strategy allows the model to regress missing information in synthesized stable frames from neighboring frames. 
For example, a temporal window of $5$ consecutive unstable frames is used in DMBVS (\ie,~$S_t=\{I_{t - 2},I_{t - 1}, I_{t}, I_{t + 1},I_{t + 2}\}$), 
and a temporal window of $3$ consecutive frames with frame recurrence is used in DIFRINT (\ie,~$S_t = \{\hat{I}_{t - 1}, I_{t}, I_{t + 1}\}$).
Notably, this approach enhances only the midframe $I_t$ within the window.

These pixel-synthesis methods are straightforward and allow for end-to-end learning and inference. However, one of the main drawbacks of these methods is their limited control over stability. While iterative schemes can enhance stability, such approaches may also compromise image quality and introduce wobble (jitter) artifacts over multiple iterations. 
Despite the limited performance in stabilization, pixel-level synthesis solutions are still promising
because they can easily produce full-frame videos after stabilization. Therefore, we formulate our fast adaptation method based on these pixel-level synthesis approaches to further improve both stability and image quality at the test stage.
Notably, our proposed algorithm is model agnostic and is capable of generalizing across various stabilization models, and can be seamlessly integrated into conventional regression-based video stabilization methods.

\subsection{Meta-learning for video stabilization} \label{main_method}
During our experimentation, we observed that pixel-synthesis models for video stabilization encounter challenges in handling motion in certain scenarios due to the bias present in the training data and difficulties in evaluating/exploiting motion cues from raw pixel values. 
These key observations underscore the challenges that pixel-level synthesis stabilization models face when dealing with motion in specific scenarios. 
Therefore, we hypothesize that in real-world videos, the motion profiles can vary significantly even within the same video, for which models with fixed network parameters might be ineffective; thus, to make these models more effective without compromising their efficiency in producing high-quality full-frame videos, we propose a fast test-time adaptation strategy that allows these models to explicitly look for and utilize visual cues for input specific scenarios for better compensation of camera shakes in a self-supervised manner. 
Specifically, to guide the adaptation process, we use MAML~\cite{maml}, which is known for its ability to adapt effectively to new tasks. 
The MAML algorithm consists of two update components: an inner loop and an outer loop updates.
Within the inner loop, the parameters of the models are adapted through a small number of adaptation steps for each specified task. 
When the inner loop is completed, the outer loop trains the model parameters that can adapt more quickly to the task-specific parameters obtained from the inner loop, and these two update steps are repeated.

In this work, to define a scene-specific video stabilization approach, we consider a short sequence of frames as a ``\emph{task}''; which is then used for fast adaptation to new unseen videos through the proposed algorithm.
We consider a feed-forward video stabilization network $f_{\theta}$, which takes a set of $\emph{2k + 1}$ neighboring frames as in Eq.~\ref{eq:frame_ip_op_non_rec} to render a stable frame $\hat{I}_{t}$. 
Our method works with conventional pixel-synthesis methods and we use DMBVS and DIFRINT as baselines.
Our task involves minimizing stability and image quality objectives within the MAML framework on local temporal sequences consisting of $q$ consecutive frames from unstable videos. 
Notably, $q = r + 2k$ and we set $r = 5$ and $k$ to be adjusted according to the selected baseline model ($k=1$ for DIFRINT and $k=2$ for DMBVS). 
The input video can be divided into $p$ consecutive non-overlapping tasks for both training and adaptation. 
Please refer to Sec.~\ref{tgt_adapt} for details of various task sampling strategies during adaptation.

\begin{figure}[t]
    \centering
    \includegraphics[width=1.0\linewidth]{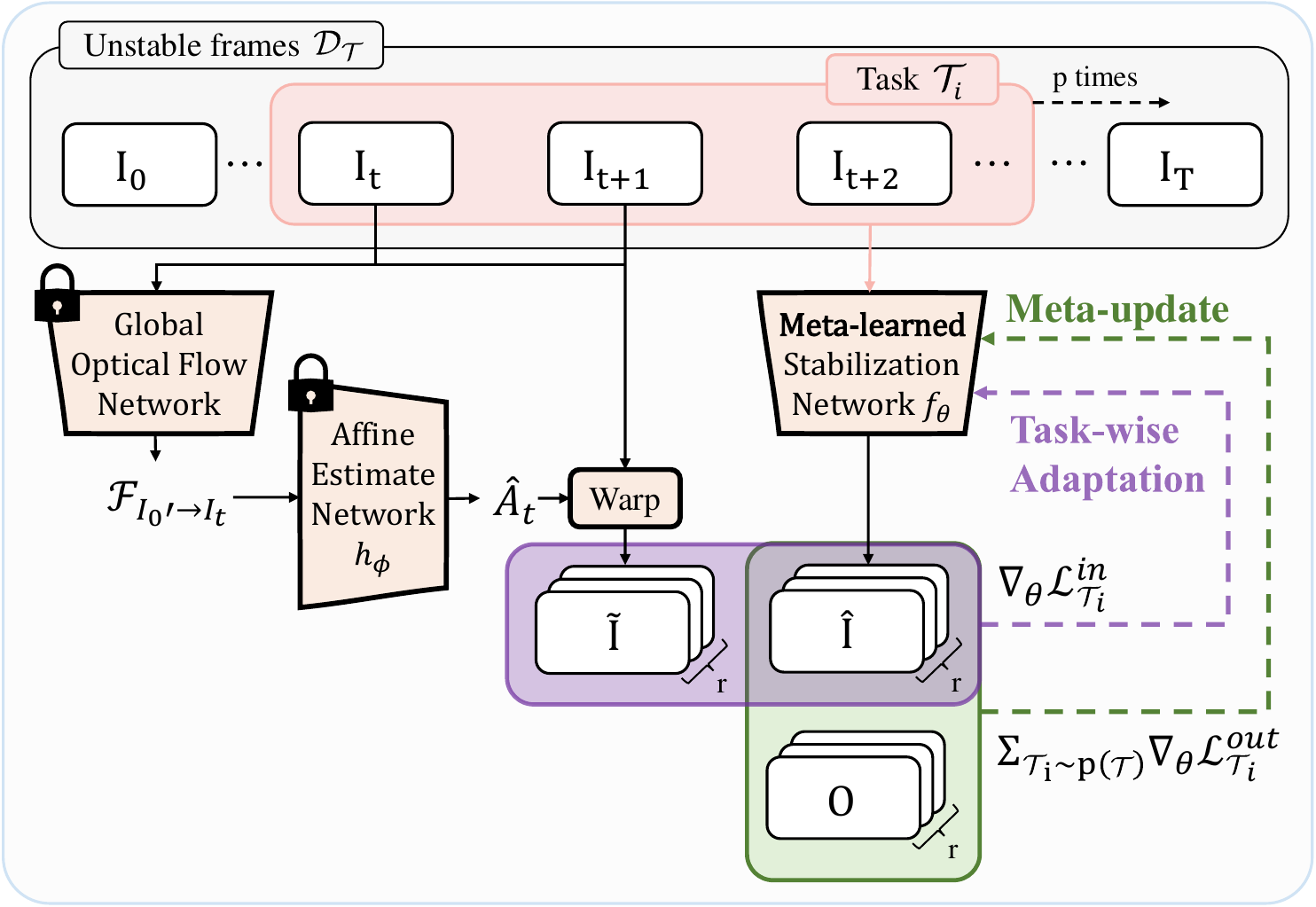}
    \caption{\textbf{Meta-training pipeline overview.} 
     The model in the inner loop gets a local sequence of $q$ consecutive frames ($\mathcal{T}_i \in \mathcal{D_{T}}$) and synthesizes corresponding stable frames.
     Inner loop penalizes synthesized frames against aligned frames; outer loop measures deviation from ground-truth stable frames.
     At inference time, only the inner loop optimization is needed. 
     Please note that $I_{0'}$ in $\mathcal{F}_{I_{0'} \rightarrow I_{t}}$ represents the initial frame of the sampled local temporal sequence, used to align the short video clips (local temporal sequences).}
    \label{overview_fig}
    \vspace{-1em}
\end{figure}

Our meta-training process is illustrated in Fig.~\ref{overview_fig}.
During the training-phase, each task $\mathcal{T}_i$ is sampled from the videos ($\mathcal{D}_{\mathcal{T}}$) in DeepStab dataset~\cite{wang2018deep}.
The inner loop update is governed with the help of an inner loop loss function $\mathcal{L}^{\text{in}}_{\mathcal{T}_i}$ which does not require the ground truth counterpart, whereas the parameter update at the meta-stage (outer loop) is governed by an outer loop loss function $\mathcal{L}^{\text{out}}_{\mathcal{T}_i}$ for which we utilize the ground-truth stable videos in DeepStab dataset. 
In our formulation, the inner loop loss optimizes for input-specific information available at test time.
In contrast, the outer loop loss prioritizes general information from the training dataset to mitigate camera shake and jerk-related artifacts such as blur; hence, it requires stable counterparts. 
Thus, meta-learning is employed to utilize both the input-specific cues available at test time and various aspects of traditional video stabilization. 
\subsubsection{Objective Functions}
Inspired by the findings of Ali~\etal~\cite{bmvc_ver}, which demonstrate the feasibility of using an image space reconstruction objective as a proxy for stabilization loss, we define our motion penalties implicitly in both the image space and the optical flow space. 

In the image space, we utilize the proposed rigid transform estimation module. This formulation addresses the absence of well-aligned ground truth data for video stabilization and corrects perspective mismatches in the DeepStab dataset, as noted in~\cite{bmvc_ver, yu2020learning}.
We specifically choose rigid affine transforms, which exclude scale and shear.
Our rigid affine transforms enhance stabilization, reduce visual distortion, and mitigate common wobble artifacts in video stabilization~\cite{wang2018deep, liu2021hybrid}.

For the rigid affine transform, we train a separate network on randomly transformed images, using an encoder-decoder architecture with a fully connected head to regress rotation and translation parameters via global optical flow $\mathcal{F}_{I \rightarrow I'}$ between two images $I$ and $I'$, as in \cite{geo2023globalflownet}.
We use global optical flow instead of conventional optical flow to mask local motion differences caused by moving objects or depth changes.
This allows the rigid transform estimation network to favor global over local transforms in videos and remains effective at the image boundaries, as shown in Fig.~\ref{fig:flow_comp}.
The proposed network regresses the rigid affine parameters as follows:
\begin{figure}[t]
    \centering
    \includegraphics[width=0.90\linewidth]{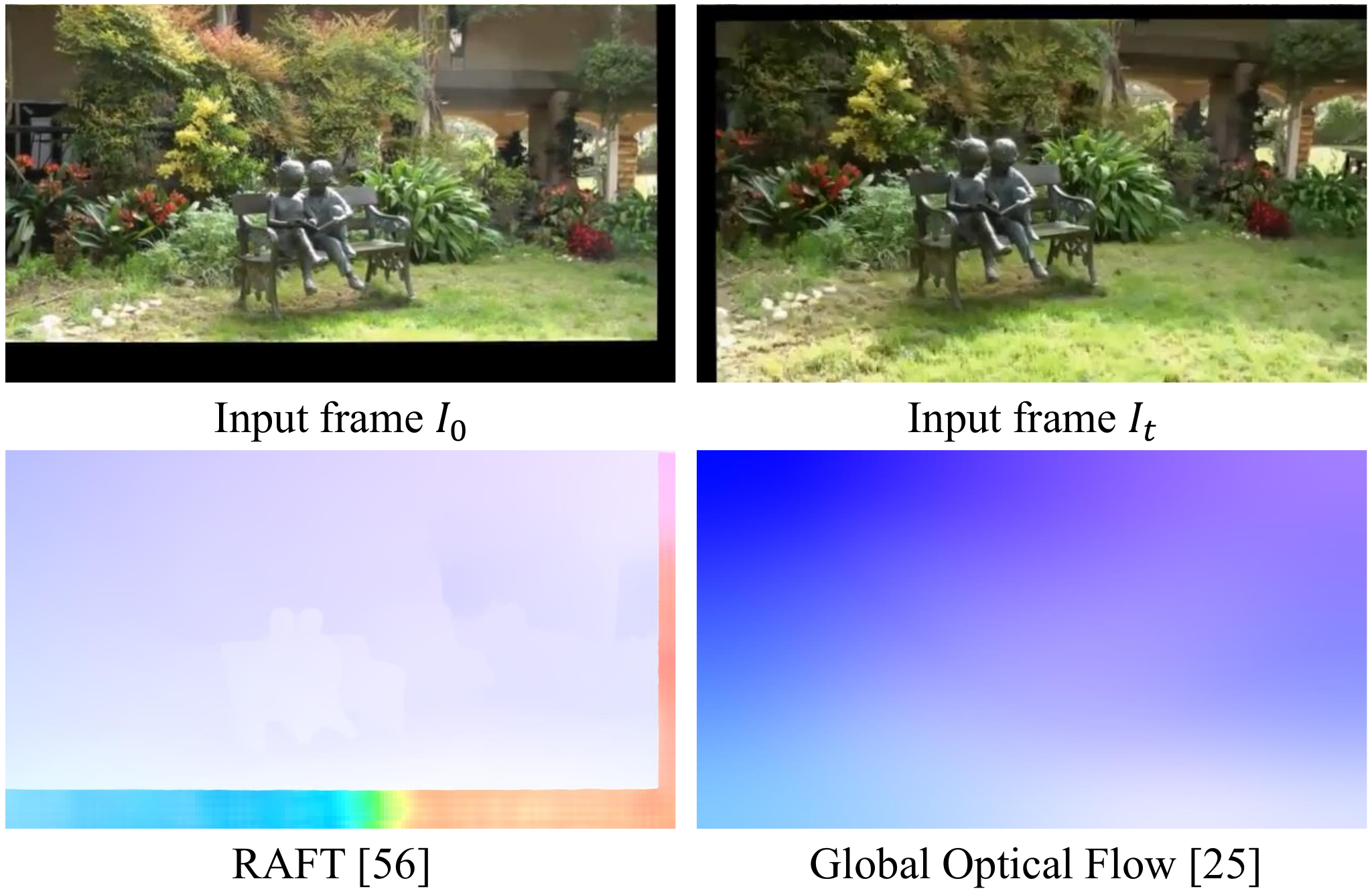}
    \caption{\textbf{Comparison of conventional and global optical flow.} Optical flow estimated from RAFT~\cite{teed2020raft} (bottom left) and Global Optical Flow~\cite{geo2023globalflownet} (bottom right). Global optical flow, in addition to masking locally different motions by dynamic objects or abrupt depth changes, also fills in the gaps near the frame boundaries introduced due to coarse alignment and only provides global motion, which supports coarse alignment.}
    \label{fig:flow_comp}
\end{figure}

\begin{equation}
\hat{\mathcal{A}} = 
\begin{bmatrix}
\cos(\theta) & -\sin(\theta)  & x \\
\sin(\theta) & \cos(\theta)   & y \\
0 & 0 & 1
\end{bmatrix}
= h_{\phi} (\mathcal{F}_{I \rightarrow I'}),
\end{equation}
where $\hat{\mathcal{A}}$ denotes the estimated rigid-affine transform, $h_{\phi}$ is the proposed affine estimation network, $\theta$ and $(x, y)$ are the rotation and translation components, respectively. 
The proposed affine estimation network is trained with an affine loss and a pixel loss defined as follows:

\begin{figure}[t]
    \centering
    \includegraphics[width=1.0\linewidth]{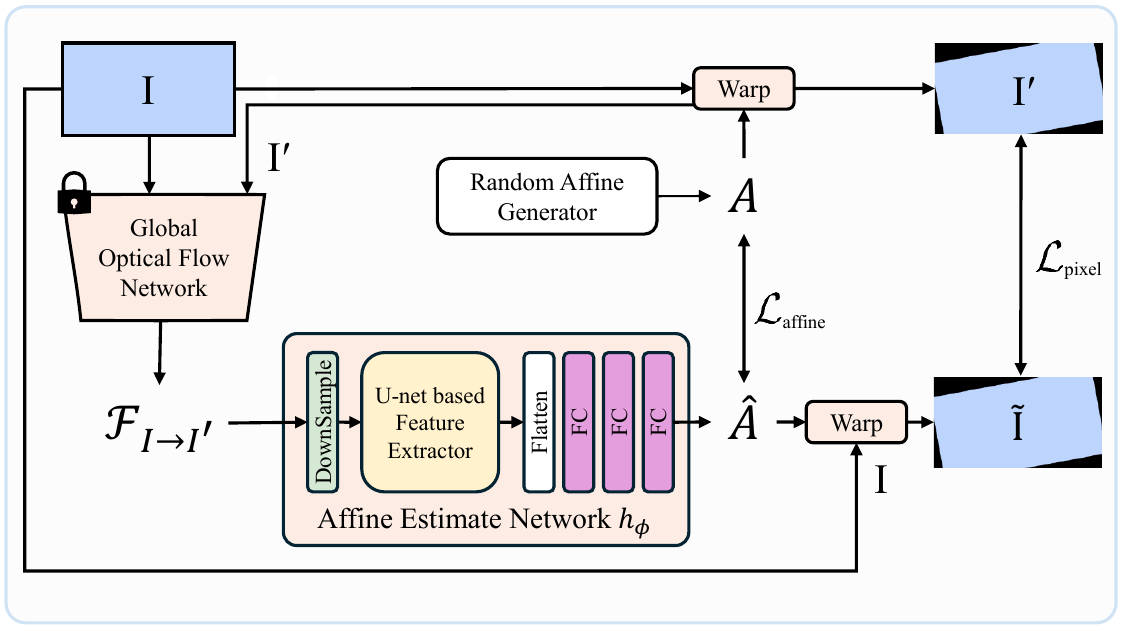}
    \caption{\textbf{Affine estimation network.} 
    Overview of architecture and the training pipeline of the proposed affine estimation network. The input image $I$ is transformed to $I'$ with a random rigid affine ${A}$, $I'$ is used for both global flow estimation and as the target for the pixel loss $\mathcal{L}_{\text{pixel}}$ to train this network. The network learns to regress rigid affine parameters $\hat{A}$, containing rotation and translation, from the estimated global flow. Once trained, it can be used to align unstable frames via rigid affine estimation.}
    \label{overview_fig_affine}
    \vspace{-1em}
\end{figure}

 \begin{equation}
 \mathcal{L}_{\text{affine}} = \left\| \mathcal{A} - \hat{\mathcal{A}}\right\|_{2}^{2},
 \label{eq:aff_eq_net}
 \end{equation}

 \begin{equation}
     \mathcal{L}_{\text{pixel}} = \left\| I - w(I', \hat{\mathcal{A}}) \right\|_{2}^{2}.
 \label{eq:pix_eq_net}
 \end{equation}
For the affine loss in Eq.~\ref{eq:aff_eq_net}, $\mathcal{A}$ denotes the ground truth rotation and translation components of the random affine transform, and for the pixel loss in Eq.~\ref{eq:pix_eq_net}, $w(\cdot)$ denotes the warping operator which warps the transformed frame $I'$ towards its untransformed counterpart $I$. Both of these losses are used in the training of the proposed rigid affine estimation network. Fig.~\ref{overview_fig_affine} illustrates the training pipeline of the proposed affine estimation network. The training of this model converged in nearly $\sim$40K optimization iterations. After convergence, this network
is used to align short sequences of input frames by estimating transformation parameters w.r.t. the first input frame as follows:
\begin{figure}[t]
    \centering
    \includegraphics[width=0.98\linewidth]{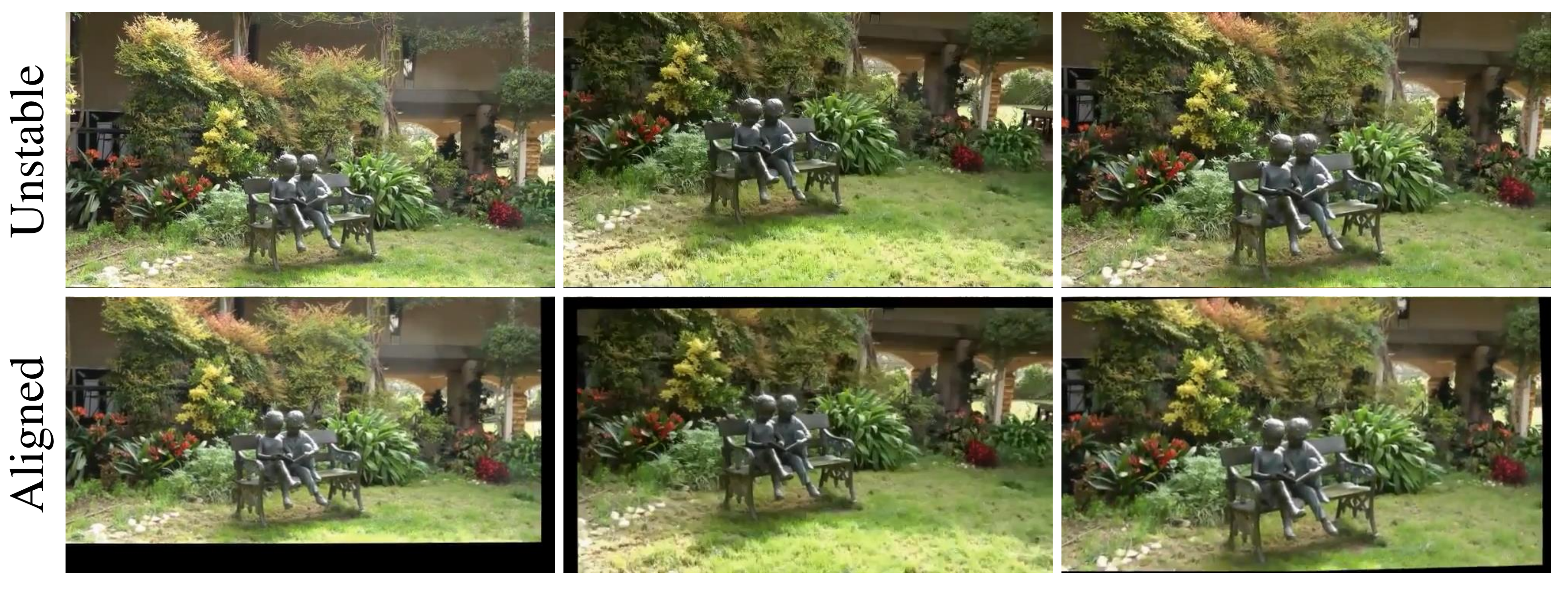}
    \caption{\textbf{Affine alignment.} This affine alignment strategy is analogous to the classical stabilization strategies which estimate and smooth transforms to stabilize videos. Please note that these frames are not neighboring frames and were selected to highlight the crops near the image boundaries in aligned frames $\tilde{V}$.}
    \label{coarse_stable_frame}
    \vspace{-1em}
\end{figure}
\begin{equation}
\begin{split}    
    &\mathcal{\hat{A}}_{t} = h_{\phi} (\mathcal{F}_{I_{0'} \rightarrow I_{t}}),~t \in \{1, ..., T'\},\\
    &\tilde{I}_{t} = \mathcal{W}(I_{t}, \mathcal{\hat{A}}_{t})~, ~\tilde{V} = \{\tilde{I}_{0'}, ..., \tilde{I}_{T'}\}.
\end{split}
\label{eq:8}
\end{equation}
Here, $\mathcal{\hat{A}}_{t}$ denotes the estimated rigid transform that maps frame $I_t$ to the first frame ($I_{0'}$) of the sequence, $T'$ denotes the number of frames in sampled short sequences, $\mathcal{W}$ represents the spatial warp operator, and $\tilde{I}_t$ refers to the warped frame, and the set $\tilde{V}$ represents the aligned frames.
Please note that ${\tilde{I}_{0'}}$ denotes the initial frame of the sampled short sequences instead of the true initial frame of the video (such that $0 \leq 0' \leq T' \leq T$). 
Notably, $I_{0'}$ serves as a reference frame, so alignment is not necessary, and $I_{0'}$ is used interchangeably with $\tilde{I}_{0'}$.
While the aligned frames provide a stabilization guide, they contain substantial boundary crops, as shown in Fig.~\ref{coarse_stable_frame}, preventing their direct use as ground-truth stable frames like in DMBVS.

Therefore, we enforce global optical flow space regularization and define stability loss as the absolute mean of global optical flow between the regressed frame $\hat{I}_t$ in ${\mathcal{T}_i}$ and the rigid-affine aligned frame $\tilde{I}_t$:
\begin{equation}
    {\mathcal{L}^{\text{in}}}_{\text{stability}} = \sum_{t = 0'}^{T'} \frac{1}{N} \sum_{N} | \mathcal{F}_{\hat{I}_t \rightarrow \tilde{I}_t} |,
\end{equation}
where $N$ represents the total number of pixels in the regressed frame. Please note that the global optical flow estimation is quite robust against the cropped regions in the warped frames $\tilde{I}_t$ and fills these holes using the visual context from the input images.
The intuition behind this loss formulation is to enforce dense alignment between the regressed and aligned sequences, as ideally the regressed frames and the aligned frames should align perfectly.
However, the stability loss alone cannot ensure clear content synthesis, due to multiple solutions to the optical flow equation~\cite{btcc_work}; thus, strong visual penalties are needed to preserve content.
We introduce these penalties in the form of perceptual loss~\cite{johnson2016perceptual}, a contextual loss, and feature-based gram matrix loss to preserve the visual content and style of input videos. Please note that throughout our experiments we use $T' = r = 5$ due to resource limitations.
The proposed quality loss to preserve video quality is defined as:
\begin{equation}
\begin{split}
    {\mathcal{L}^{\text{in}}}_{\text{quality}} = &\sum_{t=0'}^{T'}\sum_{l}\left\|\phi_{l}\left(\hat{I}_{t}\right)-\phi_{l}\left(\tilde{I}_{t}\right)\right\|^2_{2} \\
    & + \sum_{t=0'}^{T'}\sum_{l}\left\|G(\phi_{l}\left(\hat{I}_{t}\right))-G(\phi_{l}\left(\tilde{I}_{t}\right))\right\|^2_{2}\\
    & - \log(CX( \phi _ l (\hat{I}_{t})  , \phi _ l (\tilde{I}_{t}))).
\end{split}
\label{eq:in_qual_loss}
\end{equation}
The initial term represents the conventional perceptual loss, which measures the distance within the feature space, where $\phi_{l}(\cdot)$ denotes the representation of the VGG-16 network at the layer $l$ trained on the ImageNet dataset~\cite{deng2009imagenet}, and we sum up the distances up to the $relu\_4\_3$ layer. 
$G$ represents the gram matrix of features extracted from the corresponding layer $l$ and $CX(\cdot)$ represents contextual loss.
We employ perceptual and contextual losses in our formulation in line with the previous literature~\cite{bmvc_ver}, which has shown the effectiveness of these losses for video stabilization.
In particular, the addition of gram matrix loss further encourages the models to synthesize realistic frames.
Then, we define our inner loop loss within the meta-learning process as the sum of stability and quality losses between these aligned frames and the regressed frames as follows:
\begin{equation}
    \mathcal{L}^{\text{in}}_{\mathcal{T}_i} = \lambda_s ^{in}\cdot {\mathcal{L}^{\text{in}}}_{\text{stability}} + \lambda_q^{in} \cdot {\mathcal{L}^{\text{in}}}_{\text{quality}},
\label{eq:inner_loss}
\end{equation}
where $\lambda_s^{in}$ and $\lambda_q^{in}$ are associated weights for stability loss and quality loss, respectively.
Combining these losses updates the meta-learning inner loop to adapt the network parameter $\theta'_i$ from $\theta$. The inner loop is repeatable $M$ times. For meta-training, we set $\lambda_s:\lambda_p = 10:1$, and $M = 1$.

Next, within the outer loop, our network parameters are updated to minimize the different stability and quality penalties for $f_{\theta'_{i}}$ w.r.t. $\theta$ on different sampled frame sequences along with their stable counterparts from the DeepStab dataset.
In the outer loop update, we focus more on the qualitative objectives due to the availability of stable videos, which contain roughly the same content with better quality as compared to the unstable videos.
The loss in the flow space for the outer update is defined as the deviation between the global camera motion of synthesized frames and their stable counterparts as: 
\begin{equation}
    {\mathcal{L}^{\text{out}}}_{\text{stability}} = \sum_{t = 0'}^{T'} \frac{1}{N} \sum_{N} \left\| \mathcal{F}_{\hat{I}_t \rightarrow \hat{I}_{t + 1}} - \mathcal{F}_{O_t \rightarrow {O}_{t + 1}} \right\|^2_{2},
\end{equation}
where $O_t$ represents the target stable frame in the DeepStab dataset corresponding to the predicted stable frame $\hat{I}_t$.
This loss further enforces the learned stability of the model under consideration with smooth real-world trajectories. Similarly to the stability loss in the inner loop, this loss alone cannot justify the preservation of legible content; therefore, a qualitative penalty is also added in the outer loop update.
To be specific, both stable and unstable videos in the DeepStab dataset contain large disjoint perspectives~\cite{bmvc_ver, yu2020learning}, a non-local criterion is needed for quality guidance. 
We take inspiration from Ali~\etal~\cite{bmvc_ver} to define our non-local quality penalty using contextual loss~\cite{mechrez2018contextual}, which compares unaligned image regions with similar semantics and has been shown to be effective in improving the quality of synthesized stable frames.
The quality loss in the outer loop with the ground-truth target $O_t$ is defined as:
\begin{equation}
    {\mathcal{L}^{\text{out}}}_{\text{quality}} = -\log(CX( \phi ^ l ({\hat{I}}_{t})  , \phi ^ l ({O}_{t}))),
\end{equation}
and the final loss for the outer update is defined as:
\begin{equation}
   \mathcal{L}^{\text{out}}_{\mathcal{T}_i} = \lambda_s^{out} \cdot {\mathcal{L}^{\text{out}}}_{\text{stability}} 
 + \lambda_q^{out} \cdot {\mathcal{L}^{\text{out}}}_{\text{quality}}.
 \label{eq:outer_loss}
\end{equation}
Here, $\lambda_s^{out}$ and $\lambda_q^{out}$ denote the weights of the stability and quality losses, respectively. To emphasize quality during the outer loop update, we empirically found that setting their ratio to 1:10 yields the most effective results.

The overall training algorithm is presented in Alg.~\ref{alg:meta_training}. It is important to note that during testing, solely the inner loop loss is necessary for the adjustment of meta-trained parameters. Subsequently, these refined parameters are employed to synthesize the concluding stabilized results in a feed-forward manner as shown in Alg.~\ref{alg:meta_inference}.


\begin{algorithm}[htbp]

\small
    \DontPrintSemicolon
    \caption{Meta-Training.}
    \label{alg:meta_training}
    \SetKwInOut{Require}{Require}
    
    \Require{task distribution $p(\mathcal{T})$, adaptation steps $M$, inner learning rate $\alpha$, outer learning rate $\beta$, window radius $k$, the number of windows $r$ }
Initialize model parameters $\theta$ \;
$q \leftarrow r + 2k$ \; 
\While{not converged}{
    
    Sample a batch of tasks $\mathcal{T}_i \sim p(\mathcal{T})$ \;
    \ForEach{task $\mathcal{T}_i$}{
        Sample $q$-length clips from $\mathcal{T}_i$\;
        Construct $r$ temporal windows $\{S_0, S_1, ..., S_{r-1}\}$\;
        $\theta_i \leftarrow \theta$\;
        \For{$m \gets 1$ \KwTo $M$}{
            Compute regressed sequence $\hat{V}$ and aligned sequence $\tilde{V}$ with $f_{\theta_{i}}$ via Eqs.~(1), (5) \;
            Compute inner loss $\mathcal{L}^{\text{in}}_{\mathcal{T}_i}$ via Eq.~\eqref{eq:inner_loss} \;
            Update $\theta_i \leftarrow \theta_i - \alpha \nabla_{\theta_i} \mathcal{L}^{\text{in}}_{\mathcal{T}_i}$ \;
        }
    }
    Sample $\mathcal{D}'_{\mathcal{T}_i} = \{ (S_0,O_0), (S_1,O_1), ..., (S_{r-1},O_{r-1})\}$ for each $\mathcal{T}_i$ \; 
    Compute outer loss $\mathcal{L}^{\text{out}}_{\mathcal{T}_i}$ for each $\mathcal{T}_i$ via Eq.~\eqref{eq:outer_loss}\;
    Update meta-parameters: $\theta \leftarrow \theta - \beta \nabla_{\theta} \sum_{\mathcal{T}_i} \mathcal{L}^{\text{out}}_{\mathcal{T}_i}$ \;
}
\end{algorithm}
\vspace{-1em}



\begin{algorithm}[htbp]

    \small
    \SetAlgoLined
    \DontPrintSemicolon
    \SetKwInOut{Require}{Require}
    \Require{meta-trained model $f_{\theta}$, test sequence $\mathcal{T}$, adaptation number $M$, learning rate $\alpha$, window radius $k$, the number of windows $r$}
    \BlankLine
    
    $q \leftarrow r + 2k$ \;
    Sample $q$-length clips from $\mathcal{T}$\;
    
    Construct $r$ temporal windows $\{S_0, S_1, ..., S_{r-1}\}$ \;
    $\theta' \leftarrow \theta$\;
    \For{$m \gets 1$ to $M$} {
        Compute predictions $\hat{\mathbf{V}}$, $\tilde{\mathbf{V}}$ with $f_{\theta'}$ via Eq.~\eqref{eq:frame_ip_op_non_rec},~\eqref{eq:8}\;
        Compute inner loss $\mathcal{L}^{\text{in}}_{\mathcal{T}}$ via Eq.~\eqref{eq:inner_loss}\; 
        Update 
        $\theta' \leftarrow \theta' - \alpha\nabla_{\theta'}\mathcal{L}^{\text{in}}_{\mathcal{T}}$\;
    }
    
    Stabilize video $\hat{\mathbf{V}} = f_{\theta'}(\mathbf{V})$ with sliding window strategy using Eq. (\ref{eq:frame_ip_op_non_rec})\;
    \Return stabilized video $\hat{\mathbf{V}}$
    
    \caption{Meta-Inference.}
    \label{alg:meta_inference}
\end{algorithm}

\subsubsection{Jerk Localization and Targeted Adaptation}
\label{tgt_adapt}
To expedite the adaptation process and make the adaptation feasible for long videos, we first experimented with a fixed number of adaptation iterations on randomly sampled sequences from test videos. We observed that even with as few as 100 adaptation steps on randomly sampled sequences from the input test videos, the adapted models produced competitive results. This was primarily due to the repetition in the motion characteristics and the content of the scene.

Moreover, we found that not all frames equally impact the adaptation process, suggesting that a more selective strategy could further improve efficiency. Inspired by this, we propose a targeted adaptation approach that leverages motion-aware frame selection through a jerk-localization module with spatial sampling from informative regions within each frame. 
This strategy substantially reduces the adaptation time for long real-world videos and improves stabilization quality, achieving competitive performance with as few as 10 carefully selected adaptation iterations on the most unstable regions.


\begin{figure}[t]
    \centering
    \includegraphics[width=0.98\linewidth]{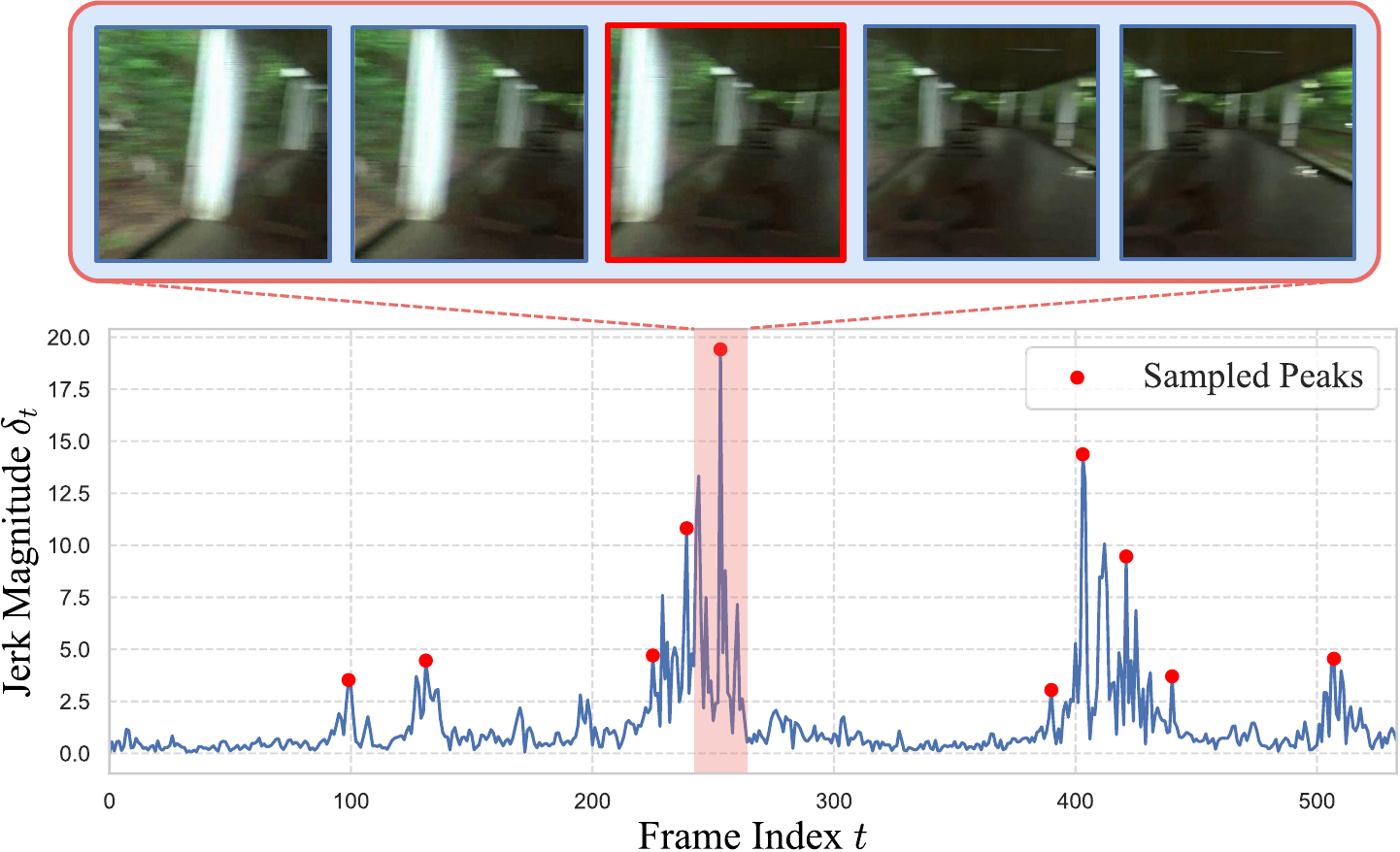}
    \caption{\textbf{Jerk Localization.} Instantaneous jerk intensity of a video from the NUS dataset is presented. Red dots mark high-magnitude, non-overlapping tasks selected for guiding the proposed targeted adaptation. For this illustration we use $p = 10$ and $k= 2$.}
    \label{fig_sampled_peaks}
    \vspace{-1em}
\end{figure}

\begin{figure}[t]
    \centering
    \includegraphics[width=0.98\linewidth]{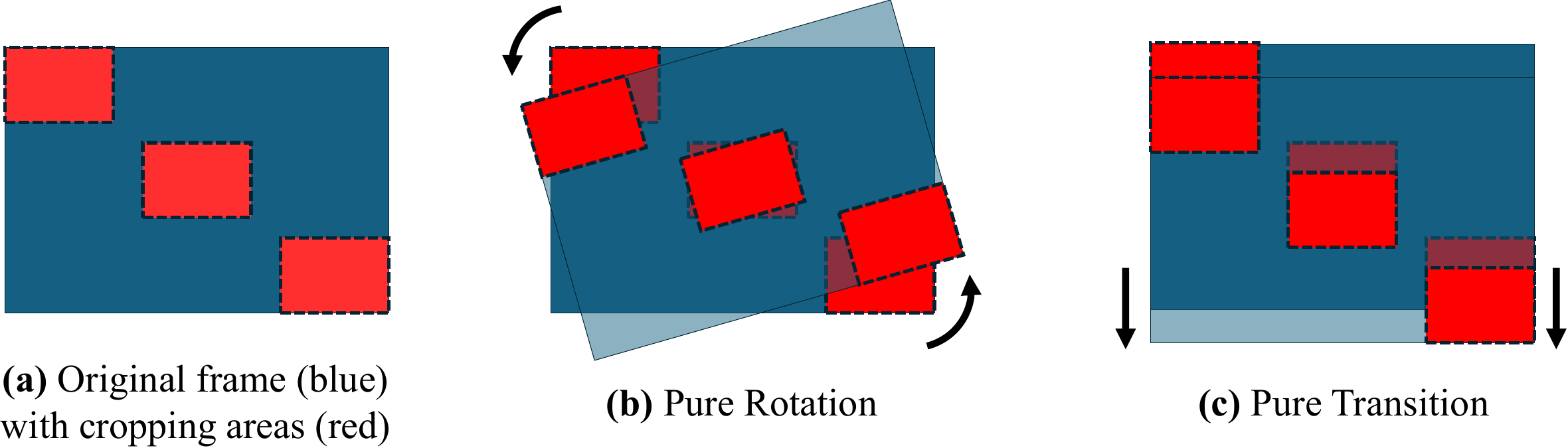}
    \caption{
    \textbf{Spatially Aware Patch Sampling Strategy.} 
Spatially aware crop selection strategy via structured crop operator $\mathcal{C}$ used for targeted optimization. 
(a) The original frame (blue) and adaptation patches selected by $\mathcal{C}$ at the corners and the center (red). 
(b) Under pure rotation, corner regions undergo the most significant displacement, making them highly informative for stability supervision, while the center patch contributes to perceptual consistency. 
(c) In pure translational motion, displacement is relatively uniform across the frame, and the selected patches provide similar supervision for both stability and quality. }
    \label{fig_spatially_aware_patch_sampling}
    \vspace{-1em}
\end{figure}

To further accelerate the meta-adaptation process, we introduce a targeted sampling strategy by identifying the most unstable regions, termed as the \textit{Jerk Localization Module}. 
Instead of uniformly adapting across entire videos in a uniform way, we aim at adapting to the most unstable segments.
These difficult unstable regions, identified by high-magnitude camera jerks, often dominate the perceived instability of the overall sequence. 
By focusing on such challenging regions, we ensure that the model quickly learns to correct the most severe jerks, which leads to improved stability across the entire video with minimal adaptation effort. 
This strategy is effective because most videos have consistent visual content and motion among frames, with varying jerk levels.
Learning from the most challenging instances enables the model to generalize effectively across videos, matching the performance of full adaptation approaches.

To detect the most unstable regions, given the estimated global motion parameters by the affine estimation network throughout the video frames, we compute the instantaneous jerk magnitude by applying the backward finite difference operator as follows:
\begin{equation}
    \Delta \mathcal{A'}_t = \mathcal{\hat{A}}_t - \mathcal{\hat{A}}_{t-1}, \quad \forall  t \in \{2, \ldots, T\},
\end{equation}
followed by computing the Euclidean norm of the motion vector as follows:

\begin{equation}
    \delta_t = \left\| \Delta \mathcal{A'}_t \right\|_2 = \sqrt{(\Delta \gamma)^2 + (\Delta x)^2 + (\Delta y)^2}.
    \label{eq:jerk_magnitude}
\end{equation}
This gives us a scalar trajectory magnitude $\delta \in \mathbb{R}^{T-1}$ representing frame-to-frame jerk intensity for all $t$.

Based on this jerk intensity, the most unstable regions of the video can be selected by applying peak detection $\boldsymbol{\delta}$ using a distance-constrained local maximum strategy following the general strategy presented in~\cite{cheng2019multiple}. Specifically, we can sample $p$ tasks centered around the highest-magnitude peaks so that no two peaks lie within a neighborhood of size $2k+1$ (temporal window size). 
These non-overlapping $p$ highest jerk tasks allow the adaptation process to see a variety of difficult content and high-motion jerks for efficiently guiding the targeted adaptation. 
In Fig.~\ref{fig_sampled_peaks}, sampled peaks are illustrated, which highlights the jerk patterns in a video from the NUS dataset~\cite{liu2013bundled} and the need to select better adaptation targets.



Building on the frame sequence selection provided by the proposed jerk localization module, we further introduce a spatially targeted adaptation strategy. During our experiments, we observed that selecting patches near the four corners and the center of the frame can lead to improved stability results. 
This aligns with motion physics: for pure rotation, displacement is greatest at frame corners; for pure translation, motion is uniform across regions.
Therefore, for rotational motion, corner regions aid in stabilization, while center regions enhance image quality.


Based on this observation, we further optimize the adaptation sampling process through structured crop placements. We empirically find that extracting patches from diagonal regions (top-left corner, center and bottom-right corner of the frame or the other diagonal as illustrated in Fig.~\ref{fig_spatially_aware_patch_sampling}) provides sufficient coverage of peripheral and central motion and content characteristics. 
This spatial sampling approach balances motion diversity and computational efficiency, allowing the model to generalize over diverse motions without extra computational cost during adaptation. 
Alg.~\ref{alg:targeted_adaptation} presents the Targeted Adaptation approach.



\begin{algorithm}[htbp]

    \caption{\small Targeted Adaptation using Jerk Localization}
    \label{alg:targeted_adaptation}
    \small
    \DontPrintSemicolon
    \SetKwInOut{Require}{Require}
    \Require{
    meta-trained model $f_{\theta}$, 
    test sequence $\mathcal{T} = \{I_0, ..., I_T\}$,
    adaptation steps $M$, learning rate $\alpha$, 
    scalar trajectory magnitude $\delta$, 
    crop operator $\mathcal{C}$,
    window radius $k$, 
    the number of windows $r$,
    the number of tasks $p$ for targeted adaptation
    }
\BlankLine


$q \leftarrow r + 2k$ \;
$\theta' \leftarrow \theta$\;
Select $p$ jerk peaks $\{j_1, ..., j_p\}$ from $\mathcal{T}$ using Eq.~\eqref{eq:jerk_magnitude} \;

\ForEach{$j \in \{j_1, ..., j_p\}$} {
    Extract $q$-length clip $C_j = \{I_j, ..., I_{j+q-1}\}$ \;
    Construct $r$ local windows $\{S_0, ..., S_{r-1}\}$ from $C_j$ \;

    \For{$m \gets 1$ to $M$} {
            Extract spatially targeted crops using $\mathcal{C}$ from Fig. (6) on each local window \;
            Compute predictions $\hat{\mathbf{V}}$, $\tilde{\mathbf{V}}$ with $f_{\theta'}$ via Eq.~\eqref{eq:frame_ip_op_non_rec},~\eqref{eq:8} \;
            Evaluate inner loss $\nabla_{\theta'} \mathcal{L}^{\text{in}}_{\mathcal{T}}$ via Eq.~\eqref{eq:inner_loss} \;
            Update $\theta' \leftarrow \theta' - \alpha \nabla_{\theta'} \mathcal{L}^{\text{in}}_{\mathcal{T}}$ \;
    }
}
Stabilize video $\hat{\mathbf{V}} = f_{\theta'}(\mathbf{V})$ using sliding window strategy \;
\Return stabilized video $\hat{\mathbf{V}}$
\end{algorithm}

Moreover, we further enhance the adaptation by adding a proxy stabilization objective, inspired by the simplistic stabilization objectives in~\cite{bmvc_ver}.
We leverage the proposed affine regression network to construct the aligned sequence and apply a photometric reconstruction loss between the aligned patches and the model outputs. 
This term encourages both motion smoothness and spatial consistency.
Although this reconstruction-based loss aids in improving performance during test-time adaptation, we also explore incorporating it into the meta-training phase. However, it causes computational overhead and substantially slow convergence while yielding minimal improvements in overall stability performance. 
Due to the resource-intensive nature of meta-training, we apply this mechanism only during the adaptation stage, where it proves beneficial.

\section{Experimental Results}
This section outlines the experimental results. 
Please see the supplementary materials for more details and results.
\subsection{Implementation Details and Evaluation Metrics}
For this work, we considered the models presented in DMBVS~\cite{bmvc_ver} and DIFRINT~\cite{DIF} as baselines. We initialized the models with the weights provided by the authors and meta-trained these models according to the training algorithm presented earlier. The dataset used in our meta-training phase was the DeepStab dataset~\cite{wang2018deep}. While the NUS~\cite{liu2013bundled}, BiT~\cite{BiT} and DOFVS~\cite{DOFVS} datasets were used for evaluation. Please note that for adaptation, both $p$ and $M$ can be adjusted according to user preference, and we discuss the effects of varying these parameters in Sec.~\ref{ablation_section}. Please refer to the supplementary material Sec. S-II for additional implementation details.

\subsubsection{Evaluation Metrics}\label{metrics}

Conventionally, the performance of Video stabilization methods is assessed using three core metrics, namely, \textit{Stability}, \textit{Cropping}, and \textit{Distortion}. Below, we present the details for evaluating each of these metrics.

\noindent\textbf{Stability:}
This metric measures the stability of a video by analyzing its representation in the frequency domain.
A one-dimensional discrete Fourier Transform is applied to the feature trajectories on $\hat{V}$, which includes both translation and rotation:
\begin{equation}
f_v = FFT(\hat{V}),
\end{equation}
where $f_v$ denotes the transformed signal. After discarding the DC component, the stability score is computed by quantifying the energy present in the $6$ lower frequency bands, following the methodology of~\cite{wang2018deep}:
\begin{equation}
S = \sum\limits_{n=2}^6 f_v(n)/\sum\limits_{n=2}^N f_v(n),
\end{equation}
where $N$ is the total number of frequency components. A higher value of $S$ indicates smoother and more stable motion.

\noindent\noindent\textbf{Cropping:}
This metric evaluates the extent to which the original scene content is retained in the stabilized video. To calculate it, the homography between each output frame and its corresponding original frame is estimated. The cropping ratio is then derived as the mean of the scale components from all estimated homographies throughout the video sequence.

\noindent\textbf{Distortion:}
This metric quantifies geometric deformation introduced during stabilization through anisotropic homography between each frame pair. Specifically, the distortion is computed using the ratio of the two largest eigenvalues from the affine matrix. The minimum of these ratios across all frames is used as the final distortion score, quantifying the visual quality of the resultant videos.

\subsection{Ablation studies} \label{ablation_section}
We perform ablation studies on the proposed components to assess effectiveness. This section focuses on adaptation strategies, while the ablations presented in the 
supplementary material Sec. S-IV 
cover loss impacts and category-specific hyperparameters.\\

\subsubsection{Finetuning vs. Meta-adaptation} 
\begin{figure}[h]
    \centering
    \includegraphics[width=0.98\linewidth]{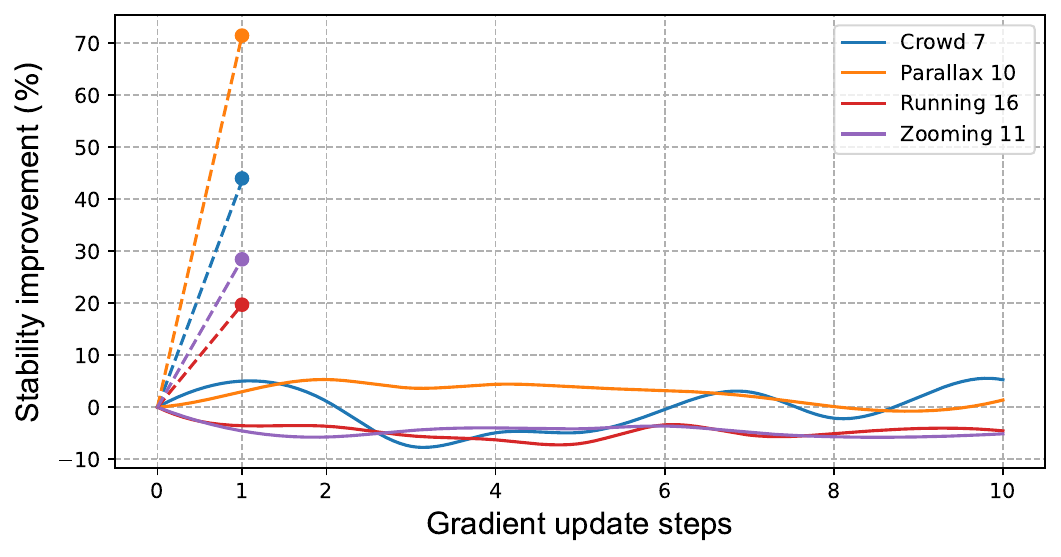}
    \\[-0.5em]
    \caption{
    \textbf{Finetuning vs. meta-adaptation.}
    The dashed lines represent the stability scores achieved by the meta-adapted models, while the solid lines denote the results obtained through naive finetuning. Naive finetuning takes more iterations for a mediocre improvement in the stability score.
    Whereas, the proposed meta-adaptation algorithm achieves a significant performance gain even with a single adaptation pass over the considered videos.
    }
    \label{fig:test_time_adaptation_experiment_fig}
\end{figure}

To highlight the efficacy and necessity of fast meta-adaptation, we conduct an ablation study by finetuning the pretrained DMBVS\footnote{Please note that similar phenomena were observed across both considered models. For clarity, we divide the ablation studies between the two models and present findings from one model in each study.}.
This is done using the proposed inner loop loss (\ie, Eq.~\eqref{eq:inner_loss}) on challenging videos from the NUS dataset where the baseline DMBVS performs poorly, with stability scores as low as 10 to 15\%.
Specifically, we compare the performance of this finetuned model with our meta-trained and meta-adapted variant, which undergoes only a single adaptation pass over the same videos.
The results of this comparison are presented in Fig.~\ref{fig:test_time_adaptation_experiment_fig}, where the meta-adaptation and finetuning outcomes are represented by dotted and solid lines, respectively.
As illustrated in Fig.~\ref{fig:test_time_adaptation_experiment_fig}, the performance gains from finetuning are unstable and not guaranteed.
For instance, the \textit{Running 16} and \textit{Zooming 11} videos show no improvement for 10 gradient updates.
In contrast, our meta-trained model demonstrates significant outperformance over the finetuned baseline, even with just a single adaptation step.

\subsubsection{Vanilla Adaptation vs. Targeted Adaptation} \label{Vanilla_VS_Target}
To evaluate the effectiveness of the proposed targeted adaptation strategy, we conducted a comparative study against a vanilla adaptation strategy using the DIFRINT model. 
The vanilla adaptation set-up performs test-time adaptation on randomly sampled frame sequences.
While the targeted adaptation leverages the proposed jerk-localization module to guide sequence selection and applies spatially-aware sampling to adapt the meta-trained model (as detailed in Sec.~\ref{tgt_adapt}). 

For this study, the same meta-trained checkpoints were used with different adaptation strategies on randomly sampled representative videos from each category of the NUS dataset. 
Tab.~\ref{tgt_vs_van_table} presents a comparative analysis of vanilla and targeted adaptations.
$\text{TargetedAdapt}^{(3)}_{10}$ and $\text{TargetedAdapt}^{(3)}_{100}$ indicate proposed targeted strategies with adaptation on 10 and 100 most unstable regions respectively, while $\text{VanillaAdapt}^{(5)}_{100}$ indicates adaptation on 100 randomly selected sequences for adaptation. The superscripts in all the presented methods represent the adaptation number, and the subscript denotes selected sequences for adaptation.


\begin{table*}[htbp]
  \centering
  \caption{\textbf{Comparison of Targeted and Vanilla Adaptation strategies.} Quantitative results comparing extended targeted adaptation ($\text{TargetedAdapt}^{\text{(3)}}_{100}$), targeted adaptation ($\text{TargetedAdapt}^{\text{(3)}}_{10}$), and fast vanilla adaptation ($\text{VanillaAdapt}^{\text{(5)}}_{100}$) on videos sampled from each category in the NUS dataset~\cite{liu2013bundled}. Targeted strategies achieve better stability overall, while extended targeted adaptation yields higher distortion scores, indicating marginally better perceptual quality. The subscript shows the number of sequences sampled for adaptation, and the superscript denotes the adaptation number.}
    \adjustbox{max width=0.98\textwidth}{\begin{tabular}{l|ccc|ccc|ccc}
    \toprule
          & \multicolumn{3}{c|}{$\text{TargetedAdapt}^{\text{(3)}}_{100}$} & \multicolumn{3}{c|}{$\text{TargetedAdapt}^{\text{(3)}}_{10}$} & \multicolumn{3}{c}{$\text{VanillaAdapt}^{\text{(5)}}_{100}$} \\
    Video & \multicolumn{1}{l}{Cropping $\uparrow$} & \multicolumn{1}{l}{Distortion $\uparrow$} & \multicolumn{1}{l|}{Stability $\uparrow$} & \multicolumn{1}{l}{Cropping $\uparrow$} & \multicolumn{1}{l}{Distortion $\uparrow$} & \multicolumn{1}{l|}{Stability $\uparrow$} & \multicolumn{1}{l}{Cropping $\uparrow$} & \multicolumn{1}{l}{Distortion $\uparrow$} & \multicolumn{1}{l}{Stability $\uparrow$} \\
    \midrule
    Crowd\_14 & 1.0000 & 0.9602 & 0.8827 & 1.0000 & 0.9554 & 0.8868 & 1.0000 & 0.9515 & 0.8539 \\
    Parallax\_14 & 1.0000 & 0.9775 & 0.8605 & 0.9998 & 0.9717 & 0.8629 & 0.9999 & 0.9699 & 0.8562 \\
    QuickRotation\_24 & 0.9998 & 0.8663 & 0.8904 & 1.0000 & 0.7992 & 0.9285 & 1.0000 & 0.9339 & 0.8839 \\
    Regular\_4 & 0.9988 & 0.9814 & 0.9004 & 1.0000 & 0.9118 & 0.8646 & 0.9998 & 0.9940 & 0.9000 \\
    Running\_8 & 0.9885 & 0.9208 & 0.8275 & 1.0000 & 0.8878 & 0.8038 & 1.0000 & 0.9129 & 0.7669 \\
    Zooming\_0 & 0.9972 & 0.9746 & 0.8788 & 1.0000 & 0.9687 & 0.8684 & 1.0000 & 0.9914 & 0.8632 \\
    \bottomrule
    \end{tabular}}%
  \label{tgt_vs_van_table}%
  \vspace{-1em}
\end{table*}%

\begin{table}[t] 
  \centering
  \caption{\textbf{Large-scale comparison of Targeted vs. Vanilla Adaptation.} Average performance across the entire NUS dataset is compared and Targeted adaptation shows higher stability with 96\% fewer adaptation steps, while vanilla adaptation obtains slightly better distortion scores, reflecting a slight trade-off between efficiency and perceptual quality.}
    \begin{tabular}{l|ccc}
    \toprule
    Method & Stability $\uparrow$ & Crop $\uparrow$ & Distortion $\uparrow$ \\
    \midrule
    $\text{TargetedAdapt}^{\text{(3)}}_{10}$ & 0.8620 & 0.9993 & 0.9354 \\
    $\text{VanillaAdapt}^{\text{(5)}}_{100}$ & 0.8528 & 0.9995 & 0.9596 \\
    \bottomrule
    \end{tabular}%
  \label{tgt_vs_van_large}%
  \vspace{-1em}
\end{table}%

From the results, we observe that \textit{Targeted Adaptation} consistently yields higher stability scores over vanilla one, validating the effectiveness of motion-informed selection in improving temporal coherence.
Interestingly, in challenging categories such as \textit{QuickRotation} and \textit{Zooming}, the \textit{VanillaAdapt} strategy achieves higher distortion scores, indicating slightly better visual quality in certain cases. 
We hypothesize that, in such videos, local high-motion scenarios with extreme jerk can limit quality when focusing exclusively on unstable regions. 
Expanding the scope of adaptation in these challenging cases, we observe that using a larger $p$ value (\ie, $\text{TargetedAdapt}^{(3)}_{100}$) achieves
better visual quality.

To further validate the effectiveness of the proposed targeted adaptation strategy, we conduct a large-scale comparison of vanilla and targeted adaptation across the entire NUS dataset and present our results in Tab.~\ref{tgt_vs_van_large}. 
The trends remain consistent: targeted adaptation outperforms the vanilla variant in terms of stability, while the vanilla strategy yields marginally better distortion scores. 
It is worth highlighting that the targeted adaptation performs 96\% fewer adaptation steps than the vanilla adaptation strategy, demonstrating a significant gain in efficiency without compromising effectiveness.

In Tab.~\ref{extended_adapt_table}, to further investigate the effects of the test-time adaptation with the proposed targeted strategy, we also conduct a large-scale ablation on the NUS dataset by varying peak jerk counts (\ie, $p$).
From this study, we can draw two important conclusions: 
First, using a larger $p$ value can improve the overall stability score. This suggests that additional supervision, even from regions that do not exhibit the highest motion instability, contributes positively to refining the model. Second, we observe that including regions with relatively lower jerks during extended adaptation improves distortion scores, indicating enhanced visual quality. This suggests that while high-jerk regions are more crucial for stabilization, low-jerk regions can provide useful guidance for enhancing the spatial fidelity in the stabilized videos.

These results highlight the effectiveness of the proposed targeted adaptation approach and its ability to balance temporal stability and visual quality. Moreover, the findings further hint at the broader potential for controlled test-time adaptation, where adaptation behavior can be steered by selecting frame regions according to desired objectives, such as maximizing structural stability, enhancing perceptual quality, or preserving specific spatial semantics. 

\begin{table}[t] 
  \centering
  \caption{\textbf{Large-scale comparison of Targeted adaptation on the NUS dataset.} Increasing the number of adapted sequences (\ie, $p$) from 10 to 100 yields a noticeable improvement in distortion and a slight gain in stability.}
    \begin{tabular}{l|ccc}
    \toprule
    Method & Stability $\uparrow$ & Crop $\uparrow$  & Distortion $\uparrow$ \\
    \midrule
    $\text{TargetedAdapt}^{\text{(3)}}_{10}$ & 0.8620 & 0.9993 & 0.9354 \\
    $\text{TargetedAdapt}^{\text{(3)}}_{100}$ & 0.8680 & 0.9995 & 0.9542 \\
    \bottomrule
    \end{tabular}%
  \label{extended_adapt_table}%
  \vspace{-1em}
\end{table}%

\subsection{Qualitative Results}
\begin{figure*}[t]
    \centering
    \includegraphics[width=0.95\linewidth]
    {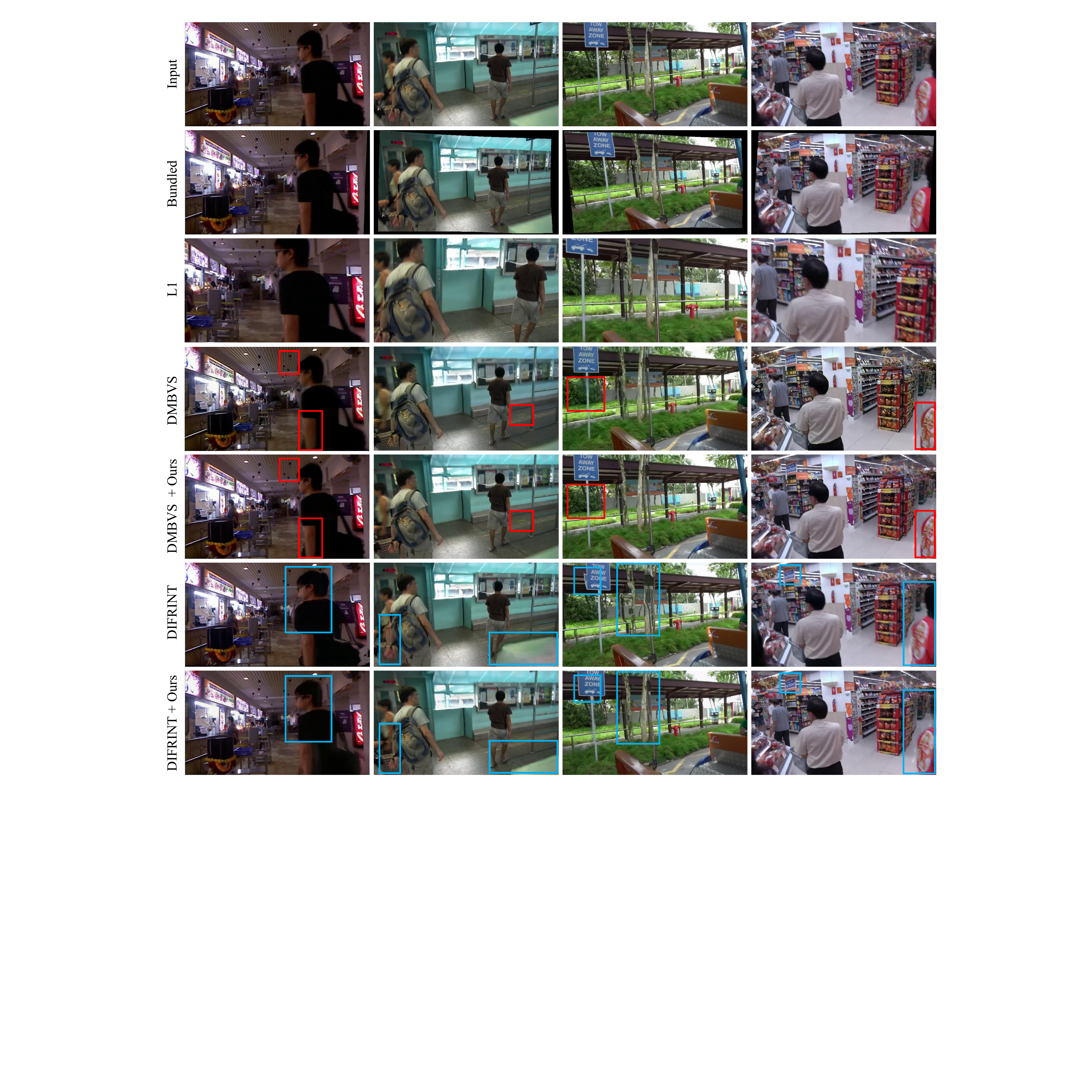}
    \caption{\textbf{Qualitative Results.} This figure presents the qualitative comparison of the meta-trained, baseline models, and classical SOTA methods. The proposed test-time adaptation strategy not only improves the stability but also enhances the visual quality of the stabilized videos (Best viewed on a computer screen with zoom).}
    \label{res:qual_comp}
\end{figure*}

For qualitative comparison, we compare our results of  ($\text{TargetedAdapt}^{\text{(1)}}_{all}$) with baseline DMBVS, DIFRINT, L1~\cite{grundmann2011auto}, Bundled~\cite{liu2013bundled}, in Fig.~\ref{res:qual_comp}.
Please note that we compare against the L1 Stabilizer~\cite{grundmann2011auto} and Bundled~\cite{liu2013bundled}, as these methods represent the SOTA in terms of stability. On the other hand, the baseline methods DMBVS and DIFRINT are included due to their strong performance in producing high-quality, full-frame stabilized videos. Recent approaches, despite their intricate designs and efficacies, fall short in comparison in terms of stability score. 
The comparison with recent and professional methodologies and the comparison with targeted adaptation are presented in the accompanied supplementary video.
The bounded regions highlight the temporal artifacts present in DIFRINT and the frame recurrent extension of DMBVS.
The proposed algorithm reduces the frequency of occurrence of these temporal artifacts and produces sharper results.
Please see supplementary material for resulting videos.

\subsection{Quantitative Results} 
\label{QuantRes}
In Tab.~\ref{tab:comp_with_baselines}-~\ref{tab:extra_mets},
we compare our quantitative results with different stabilization methods: L1~\cite{grundmann2011auto}, Bundled~\cite{liu2013bundled}, recent learning-based techniques like Fusta~\cite{liu2021hybrid}, Yu and Ramamoorthi~\cite{yu2020learning}, StabNet~\cite{wang2018deep}, Zhang~\etal~\cite{ Zhang_2023_ICCV}, and the professional software, Adobe Premiere Pro 2020. 
In all the tables, subscripts indicate sampled sequence numbers for adaptation and the superscripts show adaptation number, the best results are highlighted with green, and the second best are highlighted with blue color.

\subsubsection{Quantitative Comparison on the NUS dataset}
We first compare the performance of the test-time adapted variants against their baseline (non-adaptive) variants on the NUS dataset, using conventional metrics (\eg, Stabililty, Cropping, and Distortion). 
Our adapted models show significant improvements across all metrics, as reported in Tab.~\ref{tab:comp_with_baselines}, Tab.~\ref{crop_table}, and Tab.~\ref{dis_table}. 
Notably, the adapted version of DMBVS shows an average increase of $5\%$ in absolute stability, whereas DIFRINT achieves an average improvement of $8\%$.
It is worth mentioning that these gains do not come at the cost of compromising the full-frame nature of the models considered. Furthermore, improvement in distortion scores also highlights improved visual quality.

\begin{table}[t]
\caption{\textbf{Quantitative comparison of Stability.}
The proposed algorithm consistently improves the stability with the increasing number of adaptation iterations.
}
\adjustbox{max width=0.48\textwidth}{
  \centering
  
    \begin{tabular}{c|c|cccccc}
    \toprule
    \multirow{2}[3]{*}{Model} & \multirow{2}[3]{*}{Setting} & \multicolumn{6}{c}{Stability} \\
\cmidrule{3-8}          &       & \multicolumn{1}{c}{Crowd} & \multicolumn{1}{c}{Parallax} & \multicolumn{1}{c}{Regular} & \multicolumn{1}{c}{Running} & \multicolumn{1}{c}{Quick Rot} & \multicolumn{1}{c}{Zoom} \\
    \midrule
    \multirow{4}[1]{*}{DMBVS} & Baseline & 0.7316 & 0.7660 & 0.6939 & 0.6523 & 0.8453 & 0.7812 \\
          & $\text{VanillaAdapt}^{\text{(1)}}_{100}$ & 0.7585 & \cellcolor[rgb]{ .608,  .761,  .902}0.7966 & 0.7134 & 0.6983 & 0.8907 & \cellcolor[rgb]{ .608,  .761,  .902}0.8368 \\
          & $\text{VanillaAdapt}^{\text{(5)}}_{100}$ & \cellcolor[rgb]{ .608,  .761,  .902}0.7616 & \cellcolor[rgb]{ .663,  .816,  .557}0.8126 & \cellcolor[rgb]{ .608,  .761,  .902}0.7291 & \cellcolor[rgb]{ .608,  .761,  .902}0.7144 & \cellcolor[rgb]{ .608,  .761,  .902}0.9046 & \cellcolor[rgb]{ .663,  .816,  .557}0.8413 \\
          & $\text{TargetedAdapt}^{\text{(3)}}_{10}$  & \cellcolor[rgb]{ .663,  .816,  .557}0.7741 & 0.7924 & \cellcolor[rgb]{ .663,  .816,  .557}0.7304 & \cellcolor[rgb]{ .663,  .816,  .557}0.7184 & \cellcolor[rgb]{ .663,  .816,  .557}0.9146 & 0.7906 \\
    \midrule
    \multirow{4}[3]{*}{DIFRINT} & Baseline & 0.7453 & 0.8322 & 0.6371 & 0.7143 & 0.9059 & 0.8258 \\
          & $\text{VanillaAdapt}^{\text{(1)}}_{100}$ & 0.8062 & \cellcolor[rgb]{ .608,  .761,  .902}0.8492 & 0.6502 & 0.7218 & \cellcolor[rgb]{ .608,  .761,  .902}0.9361 & 0.8502 \\
          & $\text{VanillaAdapt}^{\text{(5)}}_{100}$ & \cellcolor[rgb]{ .608,  .761,  .902}0.8149 & \cellcolor[rgb]{ .663,  .816,  .557}0.8543 & \cellcolor[rgb]{ .608,  .761,  .902}0.6618 & \cellcolor[rgb]{ .608,  .761,  .902}0.7411 & \cellcolor[rgb]{ .663,  .816,  .557}0.9431 & \cellcolor[rgb]{ .608,  .761,  .902}0.8612 \\
          & $\text{TargetedAdapt}^{\text{(3)}}_{10}$ & \cellcolor[rgb]{ .663,  .816,  .557}0.8333 & \cellcolor[rgb]{ .608,  .761,  .902}0.8492 & \cellcolor[rgb]{ .663,  .816,  .557}0.6949 & \cellcolor[rgb]{ .663,  .816,  .557}0.7818 & 0.9268 & \cellcolor[rgb]{ .663,  .816,  .557}0.8769 \\
    \bottomrule
    \end{tabular}%
    }
  \label{tab:comp_with_baselines}%
\vspace{3Ex}
  \centering
  \caption{\textbf{Quantitative comparison of Cropping.} 
  This table highlights that despite consistently increasing the stability score, we see a minor decrease in the cropping value with increasing adaptation iterations.}
    \resizebox{\columnwidth}{!}{\begin{tabular}{c|c|cccccc}
    \toprule
    \multirow{2}[3]{*}{Model} & \multirow{2}[3]{*}{Setting} & \multicolumn{6}{c}{Cropping} \\
\cmidrule{3-8}          &       & \multicolumn{1}{c}{Crowd} & \multicolumn{1}{c}{Parallax} & \multicolumn{1}{c}{Regular} & \multicolumn{1}{c}{Running} & \multicolumn{1}{c}{Quick Rot} & \multicolumn{1}{c}{Zoom} \\
    \midrule
    \multirow{4}[1]{*}{DMBVS} & Baseline & \cellcolor[rgb]{ .663,  .816,  .557}0.9998 & \cellcolor[rgb]{ .663,  .816,  .557}0.9997 & \cellcolor[rgb]{ .663,  .816,  .557}0.9997 & \cellcolor[rgb]{ .663,  .816,  .557}0.9994 & \cellcolor[rgb]{ .663,  .816,  .557}0.9996 & \cellcolor[rgb]{ .663,  .816,  .557}0.9990 \\
          & $\text{VanillaAdapt}^{\text{(1)}}_{100}$ & \cellcolor[rgb]{ .608,  .761,  .902}0.9988 & \cellcolor[rgb]{ .608,  .761,  .902}0.9979 & \cellcolor[rgb]{ .608,  .761,  .902}0.9996 & \cellcolor[rgb]{ .608,  .761,  .902}0.9989 & 0.9961 & \cellcolor[rgb]{ .608,  .761,  .902}0.9985 \\
          & $\text{VanillaAdapt}^{\text{(5)}}_{100}$ & 0.9985 & 0.9964 & 0.9992 & 0.9965 & 0.9950 & 0.9978 \\
          & $\text{TargetedAdapt}^{\text{(3)}}_{10}$ & 0.9955 & 0.9947 & 0.9984 & 0.9965 & \cellcolor[rgb]{ .608,  .761,  .902}0.9969 & 0.9981 \\
    \midrule
    \multirow{4}[3]{*}{DIFRINT} & Baseline & \cellcolor[rgb]{ .663,  .816,  .557}0.9998 & 0.9989 & 0.9992 & \cellcolor[rgb]{ .663,  .816,  .557}0.9988 & \cellcolor[rgb]{ .663,  .816,  .557}0.9998 & \cellcolor[rgb]{ .663,  .816,  .557}0.9998 \\
          & $\text{VanillaAdapt}^{\text{(1)}}_{100}$ & 0.9996 & \cellcolor[rgb]{ .608,  .761,  .902}0.9996 & \cellcolor[rgb]{ .608,  .761,  .902}0.9996 & 0.9986 & 0.9989 & \cellcolor[rgb]{ .608,  .761,  .902}0.9997 \\
          & $\text{VanillaAdapt}^{\text{(5)}}_{100}$ & \cellcolor[rgb]{ .608,  .761,  .902}0.9997 & \cellcolor[rgb]{ .663,  .816,  .557}0.9997 & \cellcolor[rgb]{ .663,  .816,  .557}0.9997 & \cellcolor[rgb]{ .608,  .761,  .902}0.9987 & \cellcolor[rgb]{ .608,  .761,  .902}0.9992 & 0.9996 \\
          & $\text{TargetedAdapt}^{\text{(3)}}_{10}$ & 0.9996 & 0.9995 & 0.9995 & 0.9985 & 0.9984 & 0.9996 \\
    \bottomrule
    \end{tabular}%
    }
  \label{crop_table}%
\vspace{3Ex}

  \centering
  \caption{\textbf{Quantitative comparison of Distortion.} 
  This table highlights that despite consistently increasing the stability score, the quality of the processed videos is not compromised even with a higher number of adaptation iterations.}
    \resizebox{\columnwidth}{!}{\begin{tabular}{c|c|cccccc}
    \toprule
    \multirow{2}[3]{*}{Model} & \multirow{2}[3]{*}{Setting} & \multicolumn{6}{c}{Distortion} \\
\cmidrule{3-8}          &       & \multicolumn{1}{c}{Crowd} & \multicolumn{1}{c}{Parallax} & \multicolumn{1}{c}{Regular} & \multicolumn{1}{c}{Running} & \multicolumn{1}{c}{Quick Rot} & \multicolumn{1}{c}{Zoom} \\
    \midrule
    \multirow{4}[1]{*}{DMBVS} & Baseline & \cellcolor[rgb]{ .663,  .816,  .557}0.9794 & \cellcolor[rgb]{ .663,  .816,  .557}0.9660 & \cellcolor[rgb]{ .663,  .816,  .557}0.9738 & \cellcolor[rgb]{ .608,  .761,  .902}0.9064 & 0.8772 & 0.9109 \\
          & $\text{VanillaAdapt}^{\text{(1)}}_{100}$ & 0.9321 & 0.8723 & 0.9422 & 0.8251 & \cellcolor[rgb]{ .608,  .761,  .902}0.8977 & \cellcolor[rgb]{ .608,  .761,  .902}0.9396 \\
          & $\text{VanillaAdapt}^{\text{(5)}}_{100}$ & \cellcolor[rgb]{ .608,  .761,  .902}0.9415 & \cellcolor[rgb]{ .608,  .761,  .902}0.9370 & 0.9522 & \cellcolor[rgb]{ .663,  .816,  .557}0.9351 & \cellcolor[rgb]{ .663,  .816,  .557}0.9302 & \cellcolor[rgb]{ .663,  .816,  .557}0.9524 \\
          & $\text{TargetedAdapt}^{\text{(3)}}_{10}$ & 0.9265 & 0.9220 & \cellcolor[rgb]{ .608,  .761,  .902}0.9709 & 0.9033 & 0.7902 & 0.9203 \\
    \midrule
    \multirow{4}[3]{*}{DIFRINT} & Baseline & 0.9534 & 0.9544 & 0.9813 & 0.9109 & 0.8847 & 0.9299 \\
          & $\text{VanillaAdapt}^{\text{(1)}}_{100}$ & \cellcolor[rgb]{ .663,  .816,  .557}0.9626 & \cellcolor[rgb]{ .608,  .761,  .902}0.9615 & \cellcolor[rgb]{ .608,  .761,  .902}0.9878 & \cellcolor[rgb]{ .608,  .761,  .902}0.9521 & \cellcolor[rgb]{ .663,  .816,  .557}0.9366 & \cellcolor[rgb]{ .608,  .761,  .902}0.9542 \\
          & $\text{VanillaAdapt}^{\text{(5)}}_{100}$ & \cellcolor[rgb]{ .608,  .761,  .902}0.9616 & \cellcolor[rgb]{ .663,  .816,  .557}0.9634 & \cellcolor[rgb]{ .663,  .816,  .557}0.9884 & \cellcolor[rgb]{ .663,  .816,  .557}0.9542 & \cellcolor[rgb]{ .608,  .761,  .902}0.9271 & \cellcolor[rgb]{ .663,  .816,  .557}0.9697 \\
          & $\text{TargetedAdapt}^{\text{(3)}}_{10}$ & 0.9577 & 0.9611 & 0.9869 & 0.9504 & 0.8355 & 0.9374 \\
    \bottomrule
    \end{tabular}%
    }
  \label{dis_table}%
\end{table}%

Moreover, we present a comprehensive quantitative comparison against various video stabilization methods in Tab.~\ref{tab:comp_with_sotas}. 
These include classical SOTA approaches recognized for their strong stability performance L1~\cite{grundmann2011auto} and Bundled~\cite{liu2013bundled} (as noted in~\cite{VS_survey}), recent learning-based techniques Fusta~\cite{liu2021hybrid}, Yu and Ramamoorthi~\cite{yu2020learning}, StabNet~\cite{wang2018deep}, Zhang~\etal~\cite{ Zhang_2023_ICCV}, as well as professional software-based stabilizer, Adobe Premiere Pro 2020. Our findings are reported in Tab.~\ref{tab:comp_with_sotas}.
Please note that while the method proposed by Zhang~\etal~\cite{Zhang_2023_ICCV} produces outputs for the entire evaluation set, it suffers from severe shakes in the initial portion of each sequence. 
These shakes primarily stem from their design constraints involving minimum-latency, which introduce significant shakes across the initial frames, gradually leading to stable outputs. 
This early instability causes unreliable homography estimation, preventing stability metric computation. 
To ensure a fair comparison with other methods, we report the average metrics for their approach only on videos where the stability score can be consistently evaluated across the full sequence and highlight their results with ``*".

\begin{table}[t]
  \centering
    \caption{\textbf{Comparison Results on the NUS dataset.}
    The proposed algorithm consistently improves the stability with the increasing number of adaptation iterations for both baseline models.
    The proposed algorithm enables DIFRINT to achieve SOTA results with a single adaptation iteration over all the frame sequences in videos from the NUS dataset.
    Please note that the methods proposed in StabNet and Adobe Premiere Pro fail to stabilize some videos; therefore, their results are averaged over only the stabilized videos and are highlighted with ``*". 
    }
    \resizebox{\columnwidth}{!}{
    \begin{tabular}{l|c|c|c}
    \toprule
    Method & Stability $\uparrow$ & \multicolumn{1}{l|}{Cropping $\uparrow$} & \multicolumn{1}{l}{Distortion $\uparrow$} \\
    \midrule
    L1~\cite{grundmann2011auto}    & 0.8661 & 0.7392 & 0.9215 \\
    Bundled~\cite{liu2013bundled} & \cellcolor[rgb]{ .608,  .761,  .902}0.8750 & 0.8215 & 0.7781 \\
    Adobe Premiere Pro 2020*    & 0.8262 & 0.7432 & 0.8230 \\
    StabNet*~\cite{wang2018deep}    & 0.7422 & 0.6615 & 0.8878 \\
    Yu and Ramamoorthi~\cite{yu2020learning}    & 0.7905 & 0.8592 & 0.9105 \\
    FuSta~\cite{liu2021hybrid}    & 0.8037 & 0.9992 & 0.9642 \\
    Zhang et al.*~\cite{Zhang_2023_ICCV} & 0.7481 & 0.9592 & \cellcolor[rgb]{ .608,  .761,  .902}0.9988 \\
    DMBVS~\cite{bmvc_ver} (baseline) & 0.7372 & 0.9983 & 0.9189 \\
    DMBVS~\cite{bmvc_ver} + $\text{VanillaAdapt}^{\text{(1)}}_{100}$ & 0.7532 & 0.9974 & 0.9112 \\
    DMBVS~\cite{bmvc_ver} + $\text{VanillaAdapt}^{\text{(5)}}_{100}$ & 0.7852 & 0.9973 & 0.9461 \\
    DMBVS~\cite{bmvc_ver} + $\text{VanillaAdapt}^{\text{(1)}}_{\rm all}$ & 0.7760 & \cellcolor[rgb]{ .663,  .816,  .557}0.9999 & \cellcolor[rgb]{ .663,  .816,  .557}0.9990 \\
    DMBVS~\cite{bmvc_ver} + $\text{TargetedAdapt}^{\text{(3)}}_{10}$ + recurrent & 0.7990 & 0.9983 & 0.9558 \\
    DMBVS~\cite{bmvc_ver} + $\text{VanillaAdapt}^{\text{(1)}}_{\rm all}$ + recurrent & 0.7867 & 0.9999 & \cellcolor[rgb]{ .608,  .761,  .902}0.9818 \\
    DIFRINT~\cite{DIF} (baseline) & 0.7904 & 0.9993 & 0.9438 \\
    DIFRINT~\cite{DIF} + $\text{VanillaAdapt}^{\text{(1)}}_{100}$ & 0.8428 & 0.9993 & 0.9587 \\
    DIFRINT~\cite{DIF} + $\text{VanillaAdapt}^{\text{(5)}}_{100}$ & 0.8528 & \cellcolor[rgb]{ .608,  .761,  .902}0.9994 & 0.9596 \\
    DIFRINT~\cite{DIF} + $\text{TargetedAdapt}^{\text{(3)}}_{10}$ & 0.8620 & 0.9993 & 0.9354 \\
    DIFRINT~\cite{DIF} + $\text{VanillaAdapt}^{\text{(1)}}_{\rm all}$ & \cellcolor[rgb]{ .663,  .816,  .557}0.8786 & \cellcolor[rgb]{ .608,  .761,  .902}0.9994 & 0.9569 \\
    \bottomrule
    \end{tabular}}
  \label{tab:comp_with_sotas}
\vspace{-1em}
\end{table}


Overall, the proposed adaptation strategy consistently enhances the performance of both baseline models. It enables the adapted DIFRINT to attain SOTA stability score and improves the mean stability of DMBVS, while maintaining the original full-frame nature and perceptual quality of the outputs. We also extended the model proposed in DMBVS~\cite{bmvc_ver} by introducing a frame-recurrent formulation, where previously synthesized frames are fed as input in the synthesis of subsequent frames during inference. Without meta-training and adaptation, this naive extension leads to wobble-like artifacts due to error accumulation; however, the use of proposed framework mitigates these artifacts. The results of the recurrent variant of the DMBVS~\cite{bmvc_ver} model are presented with ``+ recurrent" in Tab.~\ref{tab:comp_with_sotas} -- Tab.~\ref{tab:llm-judge_1}. 
It is important to note that the overall stability performance can be further improved by increasing the number of adaptation iterations. However, due to computational constraints, the results presented in Tab.~\ref{tab:comp_with_baselines} are presented up to a single adaptation pass over the entire videos. 
Moreover, to reduce the time taken for adaptation, we also experimented with two adaptation strategies that perform adaptation on selective sequences of the videos. 
These strategies yield significant improvements over the baseline, as demonstrated in Tab.~\ref{tab:comp_with_sotas}, and significantly reduce overall adaptation overhead and further reinforce the idea of control in the stabilization process.
Lastly, it is worth mentioning that incorporating the iterative interpolation strategy with frame skips introduced in~\cite{DIF} may offer additional gains in stability at the cost of visual quality.

\subsubsection{Additional Comparison with Classical SOTAs}
To better assess the performance of our proposed adaptation framework against longstanding SOTA methods, we extend our evaluation to two additional challenging public datasets, DOFVS~\cite{DOFVS} and BiT~\cite{BiT}, and present our findings in Tab.~\ref{comp_with_sotas_ovs} and Tab.~\ref{comp_with_sotas_bit}, respectively. These datasets present a diverse range of content and motion characteristics that differ significantly from the NUS dataset. Through these experiments, we make two key observations.

\begin{table}[t] 
  \centering
  \caption{\textbf{Quantitative Results on the DOFVS Dataset.}
Comparison of classical SOTAs, and adapted stabilization methods on the DOFVS dataset. 
The proposed targeted adaptation strategy significantly improves the stability of DMBVS and DIFRINT, achieving state-of-the-art performance in challenging high-resolution, low-light scenarios. 
Notably, DMBVS benefits from strong domain alignment with DOFVS due to motion similarities with the DeepStab training set, resulting in superior generalization and stabilization performance.}
    \resizebox{\columnwidth}{!}{\begin{tabular}{l|c|c|c}
    \toprule
    Method & Stability $\uparrow$ & Cropping $\uparrow$ & Distortion $\uparrow$ \\
    \midrule
    L1~\cite{grundmann2011auto}    & \cellcolor[rgb]{ .608,  .761,  .902}0.8286 & 0.7449 & 0.9231 \\
    Bundled~\cite{liu2013bundled}* & 0.8219 & 0.8240 & 0.8736 \\
    DMBVS~\cite{bmvc_ver} + recurrent (baseline) & 0.7577 & 0.9974& 0.9531 \\
    DMBVS~\cite{bmvc_ver} + $\text{TargetedAdapt}^{\text{(3)}}_{10}$ + recurrent & 0.8253 & 0.9971 & 0.9061 \\
    DIFRINT~\cite{DIF} (baseline) & 0.7481 & \cellcolor[rgb]{ .663,  .816,  .557}0.9997 & \cellcolor[rgb]{ .608,  .761,  .902}0.9559 \\
    DIFRINT~\cite{DIF} + $\text{TargetedAdapt}^{\text{(3)}}_{10}$ & \cellcolor[rgb]{ .663,  .816,  .557}0.8331 & \cellcolor[rgb]{ .608,  .761,  .902}0.9990 & \cellcolor[rgb]{ .663,  .816,  .557}0.9820 \\
    \bottomrule
    \end{tabular}}%
  \label{comp_with_sotas_ovs}%
  \vspace{-1.5em}
\end{table}%

\begin{table}[t] 
  \centering
  \caption{\textbf{Comparison Results on the BiT Dataset.}
The proposed targeted adaptation framework consistently enhances both DMBVS and DIFRINT in terms of stability and distortion on the BiT dataset. 
Notably, our proposed adaptation allows the models to produce competitive results while maintaining the full-frame nature of the baseline methods.}
    \resizebox{\columnwidth}{!}{\begin{tabular}{l|c|c|c}
    \toprule
    Method & Stability $\uparrow$ & Cropping $\uparrow$ & Distortion $\uparrow$ \\
    \midrule
    L1~\cite{grundmann2011auto}    & \cellcolor[rgb]{ .663,  .816,  .557}0.8073 & 0.7423 & 0.9113 \\
    Bundled~\cite{liu2013bundled} & 0.7909 & 0.8196 & 0.8550 \\
    DMBVS~\cite{bmvc_ver} + recurrent (baseline) & 0.7590 & 0.9915 & 0.8784 \\
    DMBVS~\cite{bmvc_ver} + $\text{TargetedAdapt}^{\text{(3)}}_{10}$ + recurrent & 0.7875 & \cellcolor[rgb]{ .663,  .816,  .557}0.9965 & \cellcolor[rgb]{ .663,  .816,  .557}0.9336 \\
    DIFRINT~\cite{DIF} (baseline) & 0.7587 & 0.9933 & \cellcolor[rgb]{.608,  .761,  .902}0.9271 \\
    DIFRINT~\cite{DIF} + $\text{TargetedAdapt}^{\text{(3)}}_{10}$ & \cellcolor[rgb]{.608,  .761,  .902}0.7962 & \cellcolor[rgb]{.608,  .761,  .902}0.9936 & 0.9157 \\
    \bottomrule
    \end{tabular}
    } 
  \label{comp_with_sotas_bit}%
\end{table}%

First, we observe a particularly significant performance gain on the DOFVS dataset, especially for the DMBVS baseline. This gain is most pronounced in challenging nighttime sequences and high-resolution scenes ($1920 \times 1080$), where existing methods often struggle. 
This is due to the strong domain alignment between DOFVS and DeepStab, utilized in the DMBVS model training and our meta-training.
Specifically, the motion profiles in the DOFVS dataset, such as the panning and walking motions that closely resemble those seen during training, thereby enabling effective generalization. 
These results further reinforce the robustness of our adaptation strategy and demonstrate its applicability across diverse and unconstrained video scenarios.

Second, these experiments reinforce the ability to improve the stability of baseline models. However, the adaptation process inherits biases and artifacts from the underlying stabilization model. For example, the DIFRINT exhibits occasional distortions near motion boundaries, which can be seen even after adaptation, but are significantly reduced compared to the baseline variant. 

\subsubsection{Proposed Metrics}

Conventional video stabilization metrics such as \textit{Stability}, \textit{Cropping}, and \textit{Distortion} are widely adopted in the literature to assess specific aspects of stabilization performance. However, these metrics frequently fail to holistically capture the temporal coherence and perceptual continuity of object motion, factors that are often better assessed through subjective user studies~\cite{bmvc_ver, DIF}. 
To bridge this gap in quantitative metrics, and to provide more objective assessment metrics for assessing stabilization quality in the broader context of applicability in automated vision systems, we propose two complementary metrics for this task. These metrics are \textit{Average Persistence} and \textit{Temporal Intersection-over-Union (Temporal IoU)}.
\textit{Average Persistence} measures how long detected objects from the first frame remain visible in subsequent frames of the stabilized video. It uses an object detector and IoU (Intersection over Union) threshold to track whether an object remains in view over time. A higher value indicates better temporal continuity of object visibility, which implies fewer disruptions from instability and cropping.
Whereas, the \textit{Temporal Intersection-over-Union (Temporal IoU)} evaluates geometric consistency by computing the IoU of detected object bounding boxes in consecutive frames. It captures inter-frame spatial alignment of objects. High temporal IoU values imply reduced jitter in objects and more stable content across time
\footnote{Please refer to the 
supplementary material Sec. S-V 
for the detailed description and implementation details of these metrics}. 
These metrics, apart from assessing the feasibility of tracking in video stabilization methods, are designed to quantify how effectively a stabilization method preserves the temporal continuity and geometric consistency of objects across frames. 
We extensively evaluate these metrics across various stabilization approaches on the NUS dataset and present our findings in Tab.~\ref{tab:extra_mets}.

\begin{table}[t] 
  \centering
  \caption{\textbf{Evaluation using the Proposed Metrics.}
Comparison of various stabilization methods using the proposed \textit{Average Persistence} and \textit{Temporal IoU} metrics on the NUS dataset. Our targeted adaptation framework achieves SOTA results on both metrics, demonstrating improved temporal coherence and object-level consistency. DIFRINT~\cite{DIF} model with targeted adaptation achieves the highest average persistence, while DMBVS~\cite{bmvc_ver} with targeted adaptation obtains the best temporal IoU. These results reflect the effectiveness of the proposed metrics in assessing not only visual smoothness but also the feasibility of object localization across frames.}
    \resizebox{\columnwidth}{!}{\begin{tabular}{l|c|c}
    \toprule
    Method & \multicolumn{1}{l|}{Avg Persistance $\uparrow$} & \multicolumn{1}{c}{Avg Temporal IoU  $\uparrow$} \\
    \midrule
    L1~\cite{grundmann2011auto}    & 17.401 & 0.5297 \\
    Bundled~\cite{liu2013bundled} & 33.434 & 0.7969 \\
    Adobe Premiere Pro 2020* & 12.711 & 0.5296 \\
    StabNet*~\cite{wang2018deep} & 10.974 & 0.5194 \\
    Yu and Ramamoorthi~\cite{yu2020learning} & 16.231 & 0.6369 \\
    FuSta~\cite{liu2021hybrid} & 34.651 & 0.8025 \\
    Zhang~\etal*~\cite{Zhang_2023_ICCV} & 32.581 & 0.7347 \\
    DMBVS~\cite{bmvc_ver} (baseline) & 33.909 & 0.7771 \\
    DMBVS~\cite{bmvc_ver} + $\text{TargetedAdapt}^{\text{(3)}}_{10}$ + recurrent & \cellcolor[rgb]{ .608,  .761,  .902}35.020 & \cellcolor[rgb]{ .663,  .816,  .557}0.8143 \\
    DIFRINT~\cite{DIF} (baseline) & 27.970 & 0.7157 \\
    DIFRINT~\cite{DIF} + $\text{TargetedAdapt}^{\text{(3)}}_{10}$ & \cellcolor[rgb]{ .663,  .816,  .557}36.254 & \cellcolor[rgb]{ .608,  .761,  .902}0.8067 \\
    \bottomrule
    \end{tabular}}%
  \label{tab:extra_mets}%
\end{table}%

In terms of the proposed metrics, our adaptation methods achieve the highest performance, which we attribute to two key design advantages: the full-frame nature of the considered stabilization methods and the better visual quality of their outputs. In contrast to cropping-based methods (e.g., L1~\cite{grundmann2011auto} and Yu~\etal~\cite{yu2020learning}), modern approaches (e.g., Fusta~\cite{liu2021hybrid} and Zhang~\etal~\cite{Zhang_2023_ICCV}) preserve the spatial context of each frame, which ensures that objects remain within view throughout the sequence. This directly improves \textit{Average Persistence}, as objects are less likely to exit the frame or become occluded due to aggressive cropping.

Furthermore, the superior spatial consistency and structural fidelity of the DMBVS output contribute to improved \textit{Temporal IoU}. 
Our meta-adaptation strategy explicitly reduces high-frequency temporal distortions, resulting in smoother frame-to-frame transitions in terms of temporal consistency. As a result, bounding boxes associated with persistent detections remain more consistent over time, even under complex motion dynamics.
Although some residual temporal artifacts inherited from the baseline network DIFRINT persist, particularly along motion boundaries, these are typically confined to regions outside the object. 
Despite this, the overall stabilization remains coherent enough to support strong detection stability, enabling our method to retain promising performance. 

Please note that the proposed metrics reflect trends consistent with established evaluation measures. This helps verify that the introduced metrics offer a fair and meaningful perspective rather than favoring specific outcomes. In this case, the proposed metrics are designed to complement existing ones and provide additional insight into temporal coherence and object-level consistency through a functional perspective. As seen in Tab.~\ref{tab:comp_with_sotas}, methods that achieve strong results across all conventional metrics (Stability, Cropping, and Distortion), like Bundled~\cite{liu2013bundled}, FuSta~\cite{liu2021hybrid}, and Zhang~\etal~\cite{Zhang_2023_ICCV} perform well under the proposed metrics (see Tab.~\ref{tab:extra_mets}). This alignment and consistency verifies that the new metrics provide broader coverage while remaining consistent with conventional standard measures. The performance of our adaptation-based models across both sets of metrics further supports the relevance and fairness of the proposed evaluation framework.


In addition to the quantitative results presented, we also provide extensive user studies to properly assess the proposed approach. We kindly refer readers to the 
supplementary material Sec. S-VI 
for additional results and user studies.

\begin{table*}[!ht]
  \centering
  \caption{Comparison of various stabilization methods using the Mean Score from LLM-as-a-Judge on the NUS dataset. This table highlights that stabilizing videos with the proposed method consistently enhances the video understanding capability across all tested LViLMs and backbone stabilization models.}
    \adjustbox{max width=0.98\textwidth}{\begin{tabular}{l|cc|cc|cc|cc|cc}
    \toprule
\multirow{2}{*}{\backslashbox{$\text{Method}$}{\kern+5.1em$\text{LViLM}$}} & \multicolumn{2}{c|}{VideoLLaVA-7B~\cite{lin2023video}} & \multicolumn{2}{c|}{ShareGPT4Video-8B~\cite{chen2024sharegpt4video}} & \multicolumn{2}{c|}{LLaVA-NeXT-Video-7B~\cite{zhang2024llavanextvideo}} & \multicolumn{2}{c|}{VideoLLaMA3-7B~\cite{zhang2025videollama}} & \multicolumn{2}{c}{Average} \\
     & Mean Score $\uparrow$ & HSR $\uparrow$ & Mean Score $\uparrow$ & HSR $\uparrow$ & Mean Score $\uparrow$ & HSR $\uparrow$ & Mean Score $\uparrow$ & HSR $\uparrow$ & Mean Score $\uparrow$ & HSR $\uparrow$ \\
    \midrule
    Unstable & 5.598 & 0.055 & 5.504 & 0.071 & 6.299 & 0.110 & 5.165 & 0.055 & 5.642 & 0.073 \\
    L1~\cite{grundmann2011auto}    & 5.299 & 0.039 & 5.291 & 0.094 & 6.047 & 0.071 & 5.268 & 0.031 & 5.476 & 0.059 \\
    Bundled~\cite{liu2013bundled} & 5.693 & 0.087 & 5.512 & 0.079 & 6.575 & 0.181 & 5.417 & 0.055 & 5.799 & 0.101 \\
    Adobe Premiere Pro 2020 & 4.181 & 0.031 & 4.157 & 0.031 & 4.906 & 0.071 & 4.543 & 0.110 & 4.447 & 0.061 \\
    StabNet~\cite{wang2018deep} & 4.591 & 0.055 & 4.260 & 0.039 & 5.134 & 0.039 & 4.181 & 0.039 & 4.542 & 0.043 \\
    Zhang~\etal~\cite{Zhang_2023_ICCV} & \cellcolor[rgb]{ .608,  .761,  .902}5.827 & 0.118 & 5.181 & 0.063 & 6.323 & 0.205 & 5.559 & \cellcolor[rgb]{ .608,  .761,  .902}0.173 & 5.723 & 0.140 \\
    Yu and Ramamoorthi~\cite{yu2020learning} & 4.984 & 0.079 & 5.008 & 0.047 & 5.717 & 0.102 & 4.772 & 0.071 & 5.120 & 0.075 \\
    FuSta~\cite{liu2021hybrid} & 5.661 & 0.063 & 5.567 & 0.087 & 6.591 & 0.197 & 5.157 & 0.079 & 5.744 & 0.107 \\
    DMBVS~\cite{bmvc_ver} (baseline) & 5.733 & 0.108 & 5.900 & 0.200 & 6.625 & 0.208 & 5.458 & 0.108 & 5.929 & 0.156 \\
    DMBVS~\cite{bmvc_ver} + $\text{VanillaAdapt}^{\text{(1)}}_{\rm all}$ + recurrent & 5.803 & 0.102 & 5.717 & 0.102 & 6.512 & 0.260 & 5.551 & 0.126 & 5.896 & 0.148 \\
    DMBVS~\cite{bmvc_ver} + $\text{TargetedAdapt}^{\text{(3)}}_{10}$ + recurrent & 5.793 & 0.132 & \cellcolor[rgb]{ .663,  .816,  .557}6.058 & \cellcolor[rgb]{ .608,  .761,  .902}0.240 & \cellcolor[rgb]{ .608,  .761,  .902}6.711 & 0.273 & 5.645 & 0.116 & 6.052 & 0.190 \\
    DIFRINT~\cite{DIF} (baseline) & 5.686 & 0.083 & 5.917 & 0.190 & 6.612 & 0.198 & 5.413 & 0.074 & 5.907 & 0.136 \\
    DIFRINT~\cite{DIF} + $\text{VanillaAdapt}^{\text{(1)}}_{\rm all}$ & \cellcolor[rgb]{ .663,  .816,  .557}6.118 & \cellcolor[rgb]{ .663,  .816,  .557}0.181 & 5.961 & 0.150 & \cellcolor[rgb]{ .663,  .816,  .557}6.850 & \cellcolor[rgb]{ .663,  .816,  .557}0.362 & \cellcolor[rgb]{ .663,  .816,  .557}6.031 & \cellcolor[rgb]{ .663,  .816,  .557}0.236 & \cellcolor[rgb]{ .663,  .816,  .557}6.240 & \cellcolor[rgb]{ .663,  .816,  .557}0.232 \\
    DIFRINT~\cite{DIF} + $\text{TargetedAdapt}^{\text{(3)}}_{10}$ & 5.708 & \cellcolor[rgb]{ .608,  .761,  .902}0.167 & \cellcolor[rgb]{ .608,  .761,  .902}6.000 & \cellcolor[rgb]{ .663,  .816,  .557}0.250 & 6.700 & \cellcolor[rgb]{ .608,  .761,  .902}0.283 & \cellcolor[rgb]{ .608,  .761,  .902}5.825 & 0.092 & \cellcolor[rgb]{ .608,  .761,  .902}6.058 & \cellcolor[rgb]{ .608,  .761,  .902}0.198 \\
    \bottomrule
    \end{tabular}}%
    \vspace{-1em}
  \label{tab:llm-judge_1}%
\end{table*}%

\subsection{Downstream Application Results}
Due to the shortcomings of conventional metrics, user studies are used as an irreplaceable measure to properly evaluate the performance of video stabilization methods.
However, these studies are subjective and irreproducible and can contain biases.
To overcome the limitations of tedious, biased, and costly human evaluations, we propose a novel reproducible method for automatically assessing video quality using Large Language Models (LLMs) as evaluators to assess video quality in a reproducible, scalable, and unbiased manner.
Specifically, we evaluate the impact of video stabilization on semantic understanding by the downstream task of video captioning.
This approach is motivated by recent studies that suggest enhancements in visual quality often translate into better performance on high-level downstream tasks~\cite{vhrn, chen2025vidcapbench}.
The goal of this experiment is to show that higher quality video stabilization improves the semantic understanding of videos by Large Video-Language Models (LViLMs).

To build this pipeline for automated assessment, we first generate video captions from unstable and stabilized videos through various approaches L1~\cite{grundmann2011auto}, Bundled~\cite{liu2013bundled}, Adobe Premier Pro 2020, StabNet~\cite{wang2018deep}, Zhang et al.~\cite{Zhang_2023_ICCV}, Yu and Ramamoorthi~\cite{yu2020learning}, Fusta~\cite{liu2021hybrid}, DMBVS~\cite{bmvc_ver}, DIFRINT~\cite{DIF}, adapted DMBVS, and adapted DIFRINT with four distinct LViLMs: Video-LLaVA-7B~\cite{lin2023video}, LLaVA-NeXT-Video-7B~\cite{zhang2024llavanextvideo}, ShareGPT4Video-8B~\cite{chen2024sharegpt4video}, and VideoLLaMA3-7B~\cite{zhang2025videollama}.
LViLM inferences were standardized with the prompt ``\textit{Describe the video.}'', using default frame sampling, max new token length for a fair comparison and in line with standard practice~\cite{zhang2025videollama, li2025vidhalluc}.
In addition, the temperature is set to $0$ with greedy search to mitigate randomness and ensure deterministic output.

Next, we use an open-vocabulary object detection model to extract fine-grained details, such as key objects, across video frames.
Since the NUS dataset lacks ground-truth captions and traditional object detection models are limited by fewer class labels, we configure OWLv2~\cite{minderer2023scaling} to detect 1,732 object categories from the union of ImageNet~\cite{deng2009imagenet} and ADE20K-847~\cite{zhou2019semantic}.
This setup ensures the capture of diverse real-world objects.
Then, detected objects serve as a proxy for upper-bound performance~\cite{fu2024blink}, supporting the evaluation task.

Finally, a separate LLM is utilized to quantify the quality of captions, adapting an LLM-as-a-Judge approach~\cite{zheng2023judging}.
This method enables efficient, unbiased, and reproducible assessment compared to costly human evaluations.
The LLM performs score-based comparison~\cite{liu2024aligning}, rating caption qualities on a 0 to 10~\cite{li2023generative, zhu2023judgelm} scale based on their alignment with the detected objects.
The output is formatted as a JSON object, including scores and brief justifications for each test set.
We select Mistral-Small-3.1 API as a judge LLM and provide the prompt  
in the supplementary material Fig. S-V.

We evaluate each stabilization method using two metrics: Mean Score and High Score Ratio (HSR).
The Mean Score is calculated as the average caption quality score for each stabilization method.
HSR represents the proportion of captions across all test cases scoring at least 8 on a 0 to 10 scale.

\begin{figure}[t]
    \centering
    \includegraphics[width=0.98\linewidth]{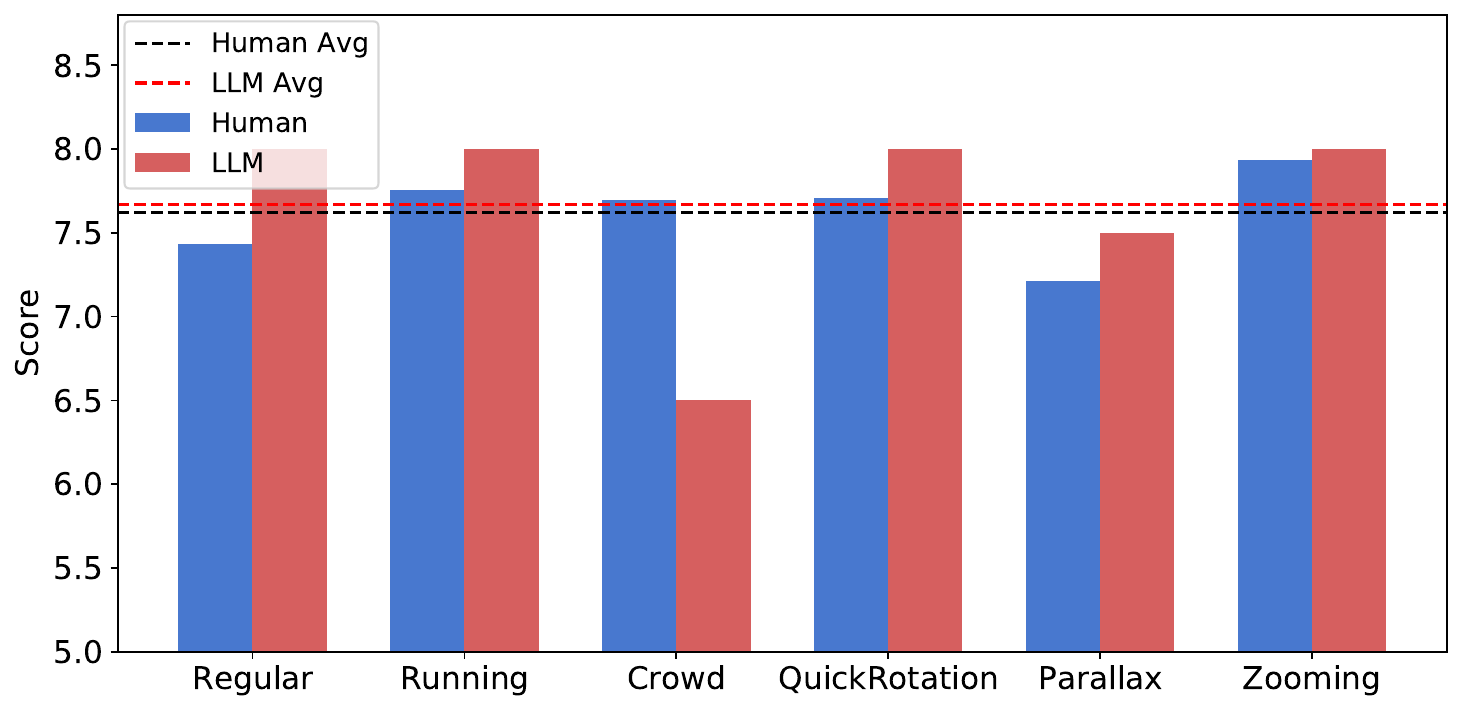} 
    \caption{ 
    \textbf{Comparison Between LLM and Human Evaluations} This figure presents the average scores by human subjects and the proposed LLM-as-a-judge pipeline. Dashed horizontal lines indicate the mean scores across all categories present in the NUS dataset. The proposed method demonstrates consistent scoring trends with human evaluations; however, a noticeable disparity appears in categories involving a large number of dynamic objects, such as Crowd from the NUS dataset. This suggests that human evaluators may be more lenient regarding completeness, whereas the LLM tends to score more strictly. 
}
    \label{llm_user_study}
    \vspace{-1em}
\end{figure}

Tab.~\ref{tab:llm-judge_1} presents the experimental results, demonstrating that the proposed adapted methods outperform not only the baselines but also the other methods in both Mean Score and HSR.
Specifically, the adapted models with DMBVS and DIFRINT have improved by 0.156 and 0.333, respectively, compared to the average Mean Score across all videos and LViLMs.
Notably, the vanilla adapted DIFRINT ($\text{VanillaAdapt}^{\text{(1)}}_{\rm all}$) achieves the highest Mean Score of 6.850 and 36.2\% of the videos achieved a score of 8 or higher by our LLM-as-a-Judge framework.
DMBVS with targeted adaptation achieves a Mean Score of 6.058 using ShareGPT4Video.
These results emphatically validate the effectiveness of our proposed adaptation method in enhancing video understanding capabilities of vision-language models and delivering superior semantic consistency for downstream tasks.
These experiments demonstrate that the proposed methodology not only enhances visual quality for human viewers but also enhances the feasibility for machine-level video understanding, which is crucial for integration with modern vision-based pipelines.
Please refer to the 
supplementary material Sec. S-VII
for the qualitative results of this experiment using VideoLLaMA3.

\subsubsection{Comparison with Human Studies}

To validate the effectiveness of the proposed LLM-as-a-Judge framework and examine the consistency between human and LLM evaluations, we conduct a user study on human preferences regarding caption quality.
For this study, we randomly sampled 12 video-caption pairs from all the categories present in the NUS dataset, and theri corresponding captions generated by LLaVA-NeXT-Video-7B.
In this user study, 31 participants were asked to rate how well the generated caption reflected the video content on a scale of 0 to 10, considering criteria such as accuracy, relevance, completeness, and clarity. 


The user study results, presented in Fig.~\ref{llm_user_study}, highlight a strong correlation between human scores and LLM-generated evaluations, confirming the reliability of the proposed LLM-as-a-Judge framework in capturing various perceptual quality aspects.
Please note that for videos with a large number of dynamic objects, such as the Crowd category of the NUS dataset, human ratings exceeded the LLM scores, suggesting that users may be more lenient or influenced by the overall content in complex scenes, leading to higher subjective scores.
However, the overall assessment shows close alignment, with the LLM averaging 7.667 and humans averaging 7.624.

The consistency observed across both human evaluators and the proposed framework confirms the practical applicability of the proposed adaptation method in modern vision-based pipelines, and it also validates the proposed evaluation framework as a suitable mechanism for assessing various video processing methods where subjective studies play a crucial role.
This evaluation approach can be extended to other tasks such as video inpainting and enhancement, where perceptual quality remains important and ground-truth supervision is often unavailable. We kindly refer the readers to the accompanied supplemental for further user studies focusing on stability, perceptual quality, and comparisons with existing stabilization methods.
\section{Conclusion}
In this work, we aim to bridge the gap between classical and modern video stabilization methods by combining their respective strengths. Our approach seeks to achieve high-quality full-frame outputs using modern deep learning techniques while incorporating the controllable characteristics typically offered by classical methods. To this end, we propose a meta-learning algorithm designed specifically for full-frame pixel-synthesis models for video stabilization. The proposed framework achieves perceptually coherent stabilization with minimal computational overhead during inference. Through the proposed targeted adaptation, the proposed framework offers practical value for real-world deployment in professional video editing tools. Extensive evaluations demonstrate that the proposed framework achieves SOTA performance on multiple real-world datasets and surpasses long-standing SOTA methods for this task. We also introduce a set of comprehensive metrics to assess the utility of video stabilization on downstream applications like tracking and object detection. The versatility of the proposed algorithm is further validated through extensive experimentation in modern video understanding pipelines. Furthermore, the proposed algorithm can be integrated seamlessly into future full-frame video stabilization pipelines without requiring additional parameters or structural modifications.


{\appendices

\bibliographystyle{IEEEtran}
\bibliography{main}


 


\begin{IEEEbiography}[{\includegraphics[width=1in,height=1.25in,clip,keepaspectratio]{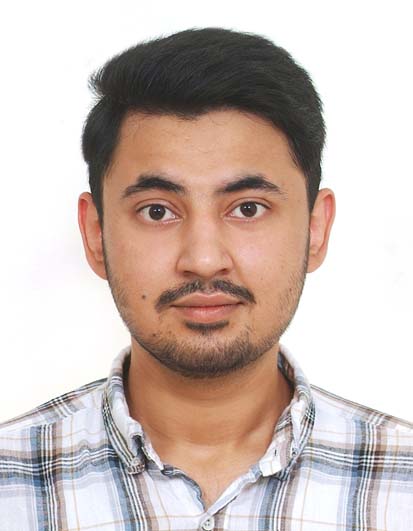}}]{Muhammad Kashif Ali} received his B.Eng degree in electrical engineering from National University of Sciences and Technology (NUST), Pakistan, and an MS-PhD degree in Computer Science from Hanyang University, Seoul, Korea, in 2018 and 2024, respectively. He is currently a postdoctoral researcher with the School of Computing and Artificial Intelligence (SCAI), Southwest Jiaotong University (SWJTU), China. His research interests include low-level computer vision and computational imaging.
\end{IEEEbiography}

\begin{IEEEbiography}[{\includegraphics[width=1in,height=1.25in,clip,keepaspectratio]{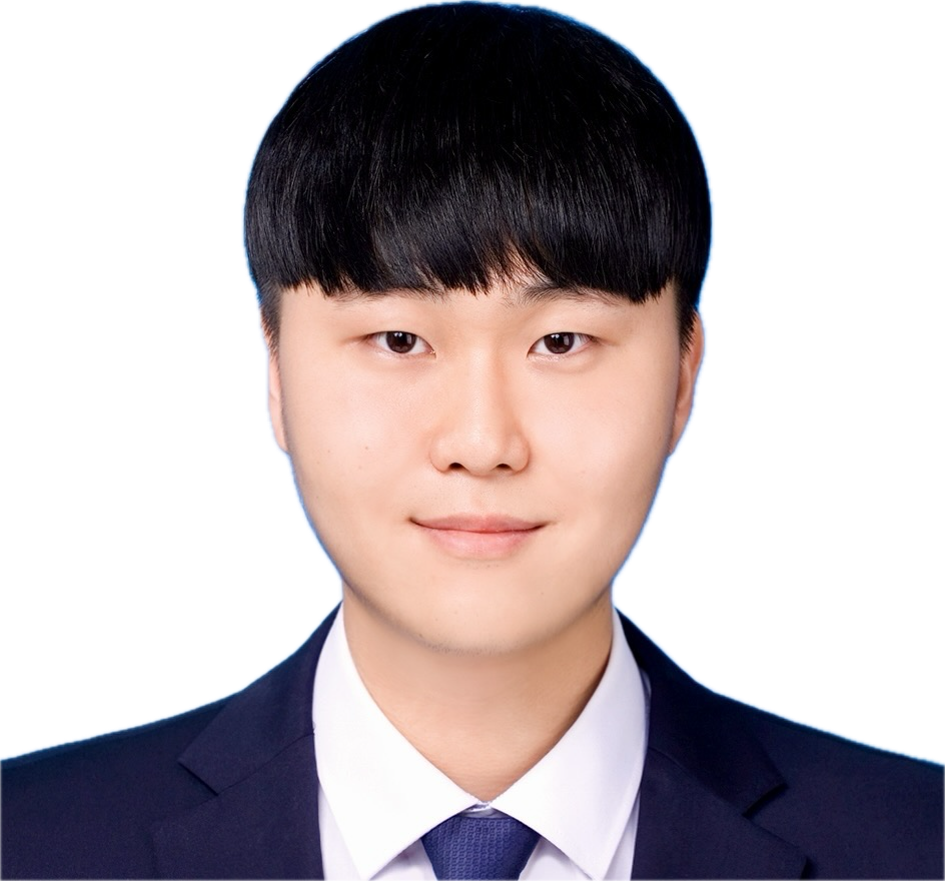}}]{Eun Woo Im} received BS degree in mathematics and an MS degree in artificial intelligence from Hanyang University, Seoul, Korea, in 2021 and 2023, respectively. He is currently a Ph.D. student in computer science at Arizona State University. His research interests include machine learning, multimodality, and their applications.
\end{IEEEbiography}
\begin{IEEEbiography} [{\includegraphics[width=1in,height=1.25in,clip,keepaspectratio]{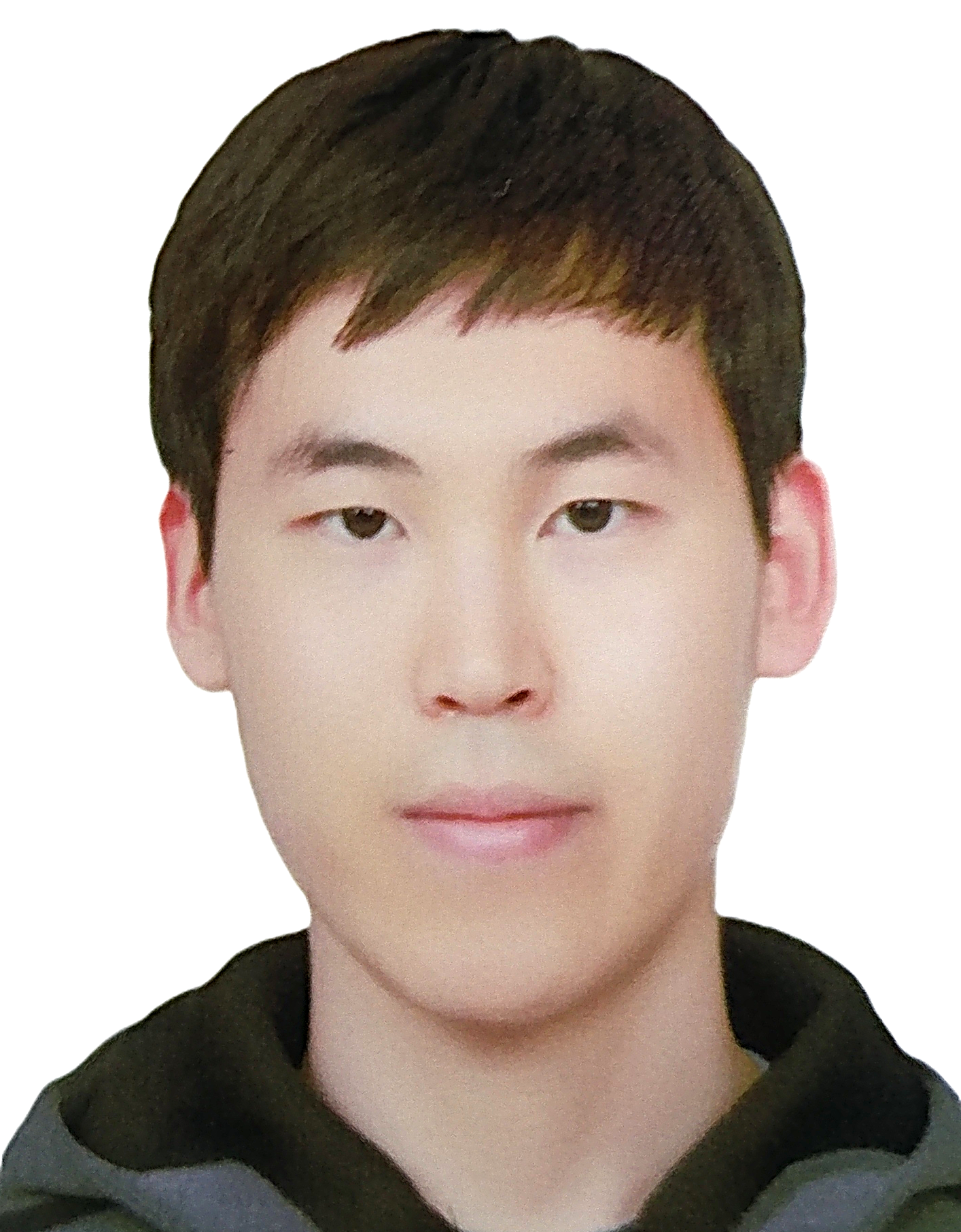}}]{Dongjin Kim} received the B.Eng degree in computer science from KonKuk University, Seoul, Korea, in 2021. He is currently an integrated MS-PhD candidate in computer science at Hanyang University, Seoul, Korea, since 2021. His research interests include low-level computer vision and computational imaging.
\end{IEEEbiography}

\begin{IEEEbiography}[{\includegraphics[width=1in,height=1.25in,clip,keepaspectratio]{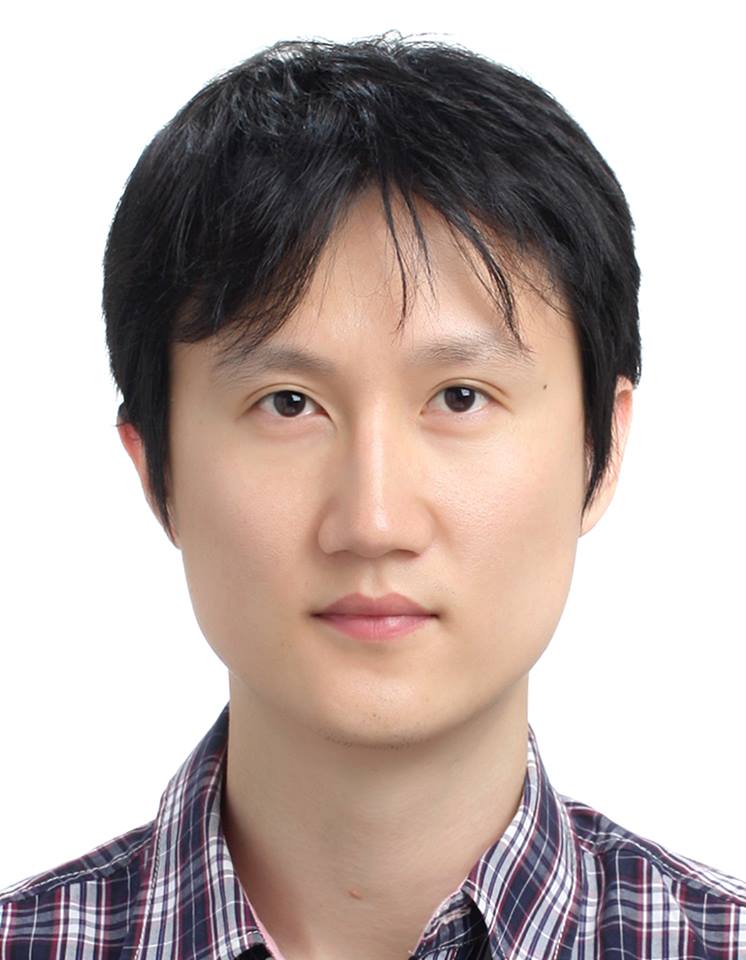}}]{Tae Hyun Kim} (Member, IEEE) received the BS and MS degrees from the Department of Electrical Engineering, KAIST, Daejeon, Korea, in 2008 and 2010, respectively, and the PhD degree in electrical and computer engineering from Seoul National University, Seoul, Korea, in 2016. He was a postdoctoral fellow with the Max Planck Institute for Intelligent Systems. He is currently an associate professor with the Department of Computer Science, Hanyang University, Seoul, Korea. His research interests include low-level computer vision and computational imaging.
\end{IEEEbiography}

\begin{IEEEbiography}[{\includegraphics[width=1in,height=1.25in,clip,keepaspectratio]{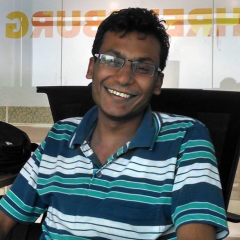}}]{Vivek Gupta} received B.Tech and M.Tech degrees in computer science and engineering from the Indian Institute of Technology (IIT), Kanpur, India in 2015 and 2016, respectively. He received the Ph.D. degree in computer science from the University of Utah. He was a postdoctoral researcher at the University of Pennsylvania. His research interests include large language model reasoning, semi-structured data, information synchronization, and knowledge integration. He is currently an assistant professor with Arizona State University.
\end{IEEEbiography}

\begin{IEEEbiography}[{\includegraphics[width=1in,height=1.25in,clip,keepaspectratio]{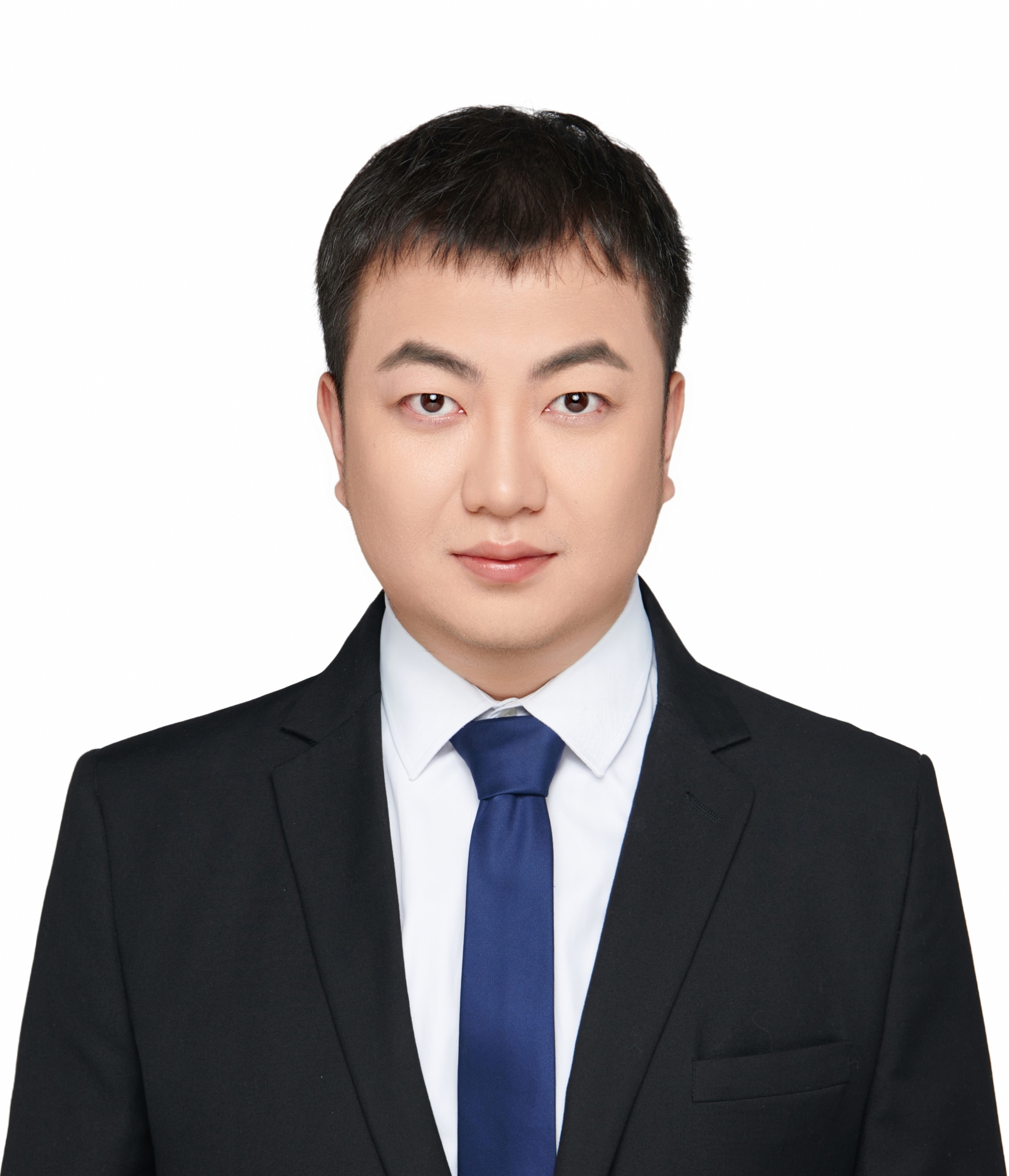}}]{Haonan Luo} received the B.Eng. degree from Taiyuan Institute of Technology, China, in 2013, and the M.S. degree from North Minzu University, China, in 2016. He received the Ph.D. degree from the School of Computer Science and Engineering, Nanjing University of Science and Technology, China. He is currently an Assistant Professor at Southwest Jiaotong University (SWJTU), China. His research interests include computer vision, embodied artificial intelligence, robotics, and reinforcement learning.
\end{IEEEbiography}

\begin{IEEEbiography}[{\includegraphics[width=1in,height=1.25in,clip,keepaspectratio]{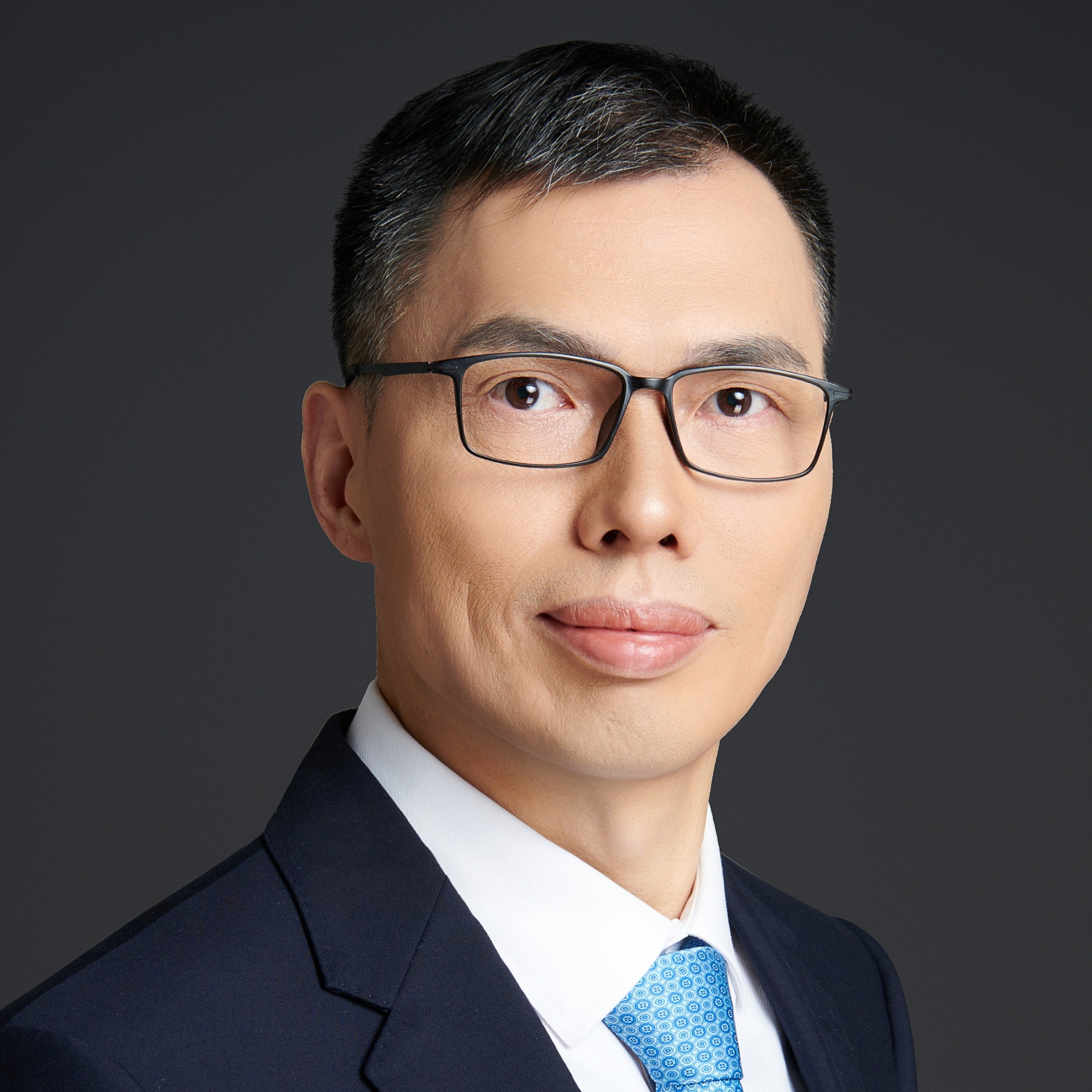}}]{Tianrui Li} (Senior Member, IEEE) received the BS, MS, and PhD degrees from Southwest Jiaotong University, Chengdu, China, in 1992, 1995, and 2002, respectively. He was a postdoctoral researcher with Belgian Nuclear Research Centre, Mol, Belgium, from 2005 to 2006, and a visiting professor with Hasselt University, Hasselt, Belgium, in 2008; University of Technology, Sydney, Australia, in 2009; and University of Regina, Regina, Canada, in 2014. He is currently a professor and the director of the Key Laboratory of Cloud Computing and Intelligent Techniques, Southwest Jiaotong University, China. He has authored or coauthored more than 300 research papers in refereed journals and conferences. His research interests include Big Data, machine learning, data mining, granular computing, and rough sets.
\end{IEEEbiography}



\vfill

\end{document}